%% file: main.tex
\documentclass[11pt]{article}

\usepackage{lmodern}

\usepackage[small,compact,center]{titlesec}

\usepackage{setspace}

\usepackage{longtable}

\usepackage{blkarray}

\usepackage[font={small,it}]{caption}

\usepackage{enumitem}\setlist{nosep}

\usepackage[letterpaper]{geometry}
\usepackage{fullpage}

\usepackage{amsthm}
\newtheorem{theorem}{Theorem}

\usepackage{color}
\usepackage{graphicx}
\usepackage{tikz}

\usetikzlibrary{arrows.meta}

\definecolor{figblue}{RGB}{0,45,245}
\definecolor{figorange}{RGB}{245, 110, 0}

\tikzset{
    vertex/.style={
        circle,
        fill=figblue,
        draw=figblue,
        inner sep=0pt,
        minimum size=7pt
    },
    solidline/.style={
        figblue,
        line width=0.8pt
    },
    dashedline/.style={
        figblue,
        line width=0.8pt,
        dashed
    },
    dottedline/.style={
        figblue,
        line width=1.1pt,
        dash pattern=on 0pt off 8pt,
        line cap=round
    },
    arrowline/.style={
        figblue,
        line width=0.9pt,
        -{Stealth[length=6pt,width=6pt]}
    },
    dashedarrow/.style={
        figblue,
        line width=0.9pt,
        dashed,
        -{Stealth[length=6pt,width=6pt]}
    }
}

\usepackage{authblk}

\usepackage{amsmath,amsfonts,amssymb,mathtools}
\allowdisplaybreaks[1]	

\usepackage[colorlinks = true,
            linkcolor = blue,
            urlcolor  = blue,
            citecolor = blue,
            anchorcolor = blue]{hyperref}

\usepackage{cleveref}

\usepackage{xr}

\usepackage{cite}

\usepackage{cancel}

\usepackage{xcolor} 

\usepackage{float} 

\usepackage{placeins} 

\usepackage{siunitx}
\usepackage{xspace}

\newcommand{\ie}{i.e.,\xspace}
\newcommand{\eg}{e.g.,\xspace}

\def\prop{^{\rm prop}} 
\def\old{^{\rm old}} 

\def\prop{_{\rm prop}} 
\def\old{_{\rm old}} 

\DeclareMathOperator{\erf}{erf}
\DeclareMathOperator{\erfc}{erfc}

\title{Bayesian methods and Markov chain Monte Carlo algorithms for curve reconstruction and point cloud data analysis}

\author[1]{Asir Intesar Tushar} 
\author[1,*]{Ioannis Sgouralis} 

\affil[1]{Department of Mathematics, University of Tennessee, Knoxville, TN, USA}

\affil[*]{Correspondence: Ioannis Sgouralis, isgoural@utk.edu}

\date{}

\usepackage{xr-hyper}
\usepackage{etoolbox}



\begin{document}

\maketitle

\begin{abstract}
Point-cloud data routinely captured by modern imaging and sensor technologies provide detailed geometric descriptions of objects and environments, but their analysis is hindered by large data volumes, localization noise, and missing information. In addition, existing point-cloud reconstruction pipelines typically return a single best-fit structure without uncertainty quantification. We introduce a fully Bayesian framework for representing point-cloud data and reconstructing closed curves, in which observed points are modeled as noisy perturbations of latent locations constrained to lie on the underlying curve that is regularized by a non-parametric prior. Posterior inference in our framework is carried out using a series of Markov chain Monte Carlo samplers tailored to point-cloud characteristics. Numerical experiments, including synthetic examples and real-world LiDAR datasets, show accurate reconstructions and quantified uncertainty over the recovered curves.
\\[1ex]
\textbf{Keywords:} point-cloud, curve reconstruction, Markov chain Monte Carlo, Bayesian methods, LiDAR
\end{abstract}

\section{Introduction}

Point-cloud data have become increasingly vital in modern technology and research due to their ability to capture detailed, multi-dimensional representations of physical objects and their surrounding environments that may range from designed to natural structures and indoor to outdoor scenes. Generated by emerging and increasingly accessible sensor technologies such as LiDAR, photogrammetry, and 3D scanners, point-clouds consist of large collections of individual localizations that collectively encode the geometry and spatial extend of solids, surfaces, and curves \cite{ritter2021robust,liu2023robust}. Point-clouds are today crucial for imaging applications across industry and academia that range from robotics \cite{chen20203d,li2020deep}, navigation systems \cite{zhang2014loam,shan2018lego}, and 3D printing \cite{akhavan2024deep} to architecture \cite{zwierzycki2016parametric}, agriculture \cite{christiansen2018ground,nidamanuri2024deep}, and ecology\cite{xi20223d,xu2023topology,price2024surface}. The ability to process and analyze point-cloud data efficiently is currently driving innovation in machine learning \cite{li2020deep,guo2020deep,sarker2024comprehensive}, computer vision \cite{fu2025consistent,luo2021diffusion}, statistics \cite{lee2022statistical,pauly2004uncertainty,senin2021statistical}, and applied mathematics \cite{hug2026minkowski,he2020curvature}.

Despite its importance for applications, the analysis of point-cloud data still faces several limitations, including high computational cost due to the large size of point-clouds, contamination with localization noise and inaccuracies caused by sensor constraints, and missing points due to environmental factors or sensor positioning, which requires processing under sophisticated mathematical models to extract meaningful insights. Furthermore, the absence of uncertainty quantification in the reconstructed structures remains an issue for applications that require accurate estimation. Addressing these challenges requires advancements in the mathematical models of point-cloud representations and also advancements in their algorithmic evaluation.

Bayesian methods are critically needed for point-cloud data analysis because they provide a robust framework to handle uncertainty, integrate prior knowledge or domain expertise, and make probabilistic inferences \cite{presse2023data,gelman2013bayesian}. Point-cloud data are often associated with significant uncertainty, and Bayesian approaches excel at quantifying and propagating this uncertainty throughout the analysis. By incorporating prior knowledge, such as expected object geometry or spatial noise patterns, and domain expertise, such as methods of data acquisition and their technological characteristics, Bayesian methods can improve the reliability of the results, even in corrupted or incomplete point-clouds. Additionally, Bayesian inference allows for adaptive modeling that may range in sophistication from simple, generic models to highly complicated, specialized ones. The probabilistic nature of Bayesian methods also facilitates decision-making under uncertainty, making them indispensable for high-stakes scenarios where analysis errors must be minimized or balanced against possible rewards \cite{presse2023data}.

In this study, we present a comprehensive, fully Bayesian method for point-cloud data analysis. Our method combines a detailed statistical model for curve reconstruction, noisy data representation, and highly efficient computational methods for its evaluation. Our novel method is not only suitable for curve reconstruction tasks in multi-dimensional point-clouds but also offers uncertain quantification on the reconstructed curves.

The remainder of this study is structured as follows. In \textsc{Methods}, we present our statistical model for point-cloud data analysis and an efficient Markov chain Monte Carlo scheme for its evaluation. In \textsc{Results}, we present characteristic demonstrations with selected examples that illustrate our method's application and overall performance in analyzing simulated as well as real-world data. Finally, in \textsc{Discussion}, we elaborate on the potential for further development and domain applications.

\section{Methods}

In this section, we first describe our mathematical model for representing point-cloud data, a novel Bayesian formulation that allows for curve reconstruction and uncertainty quantification, and finally a sequence of computational algorithms for its evaluation.

\subsection{Point-cloud representation}

We denote our point-cloud data with $w_{1:N}$. Our data consists of $d$-dimensional localizations $w_n\in\mathbb{R}^d$ that are indexed with subscripts $n=1,\dots,N$.

Each localization $w_n$ is a noisy perturbation of an idealized position $r_n\in\mathbb{R}^d$. We assume that noise perturbations are isotropic Gaussian with zero mean controlled by a parameter $\sigma>0$ that results in
\begin{align}
\label{eq:norm_like}
w_n\mid r_n,\tau &\sim \text{Normal}\left(
r_n, \sigma^2I\right), & n&=1,\dots,N.
\end{align}
Here, $\text{Normal}$ denotes a multivariate normal distribution of dimension $d$ and $I\in\mathbb{R}^{d\times d}$ the corresponding identity matrix.

The positions $r_{1:N}$ in \cref{eq:norm_like} are restricted to lie on an underlying curve $\mathcal{G}\subset\mathbb{R}^d$ that is to be reconstructed. We model $\mathcal{G}$ as a closed polyline consisting of a total of $B\ge2$ successive straight line segments which we label with superscripts $s=1,\dots,B$. Without loss of generality, we assume that each segment is directed and specified by the coordinates of its starting and ending points.

To obtain the coordinates of each segment, we consider a total of $K\gg1$ inducing points $h^k\in\mathbb{R}^d$ that we index with superscripts $k=1,\dots,K$ and use the \emph{first $B$ of them} to define the polyline. In detail, according to our convention, the polyline's $s$\textsuperscript{th} segment starts at point $h^s$ and ends at point $h^{s+1}$ for $s<B$, while the $B$\textsuperscript{th} segment starts at $h^B$ and ends at $h^1$. In this way, our curve $\mathcal{G}$ is fully specified by the geometry, which we summarize in
\begin{align*}
G=\{B,h^{1:K}\}.
\end{align*}
For each position $r_n$, we use a latent segment indicator $s_n\in\{1,2,\dots,B\}$ and a continuous local coordinate $u_n\in[0,1]$ to specify its placement along $\mathcal{G}$. Specifically, our positions are given by
\begin{align}
\label{eq:rn_impl}
r_n&=
\begin{cases}
h^{s_n}+\left(h^{s_n+1}-h^{s_n}\right)u_n, & s_n<B,
\\
h^B+\left(h^1-h^B\right)u_n, & s_n=B,
\end{cases}
 & n&=1,\dots,N.
\end{align}
With this notation, our objective from now on is the estimation of $B,h^{1:K},s_{1:N},u_{1:N}$ and $\tau$.

\subsection{Geometry representation}

A natural modeling choice for the positions $r_n$ is to assume that these are uniformly distributed along the curve
\begin{align}
\label{eq:rn}
    r_{n}&\sim \text{Uniform}_{\mathcal{G}}
    ,&
    n&=1,\dots, N.
\end{align}
This assumption reflects that, in the absence of noise, observations are equally likely to occur anywhere along the curve under reconstruction. However, working with a uniform distribution on a piecewise-linear curve such as $\mathcal{G}$ is computationally difficult. To resolve this, we introduce an alternative representation that consists of
\begin{align}
\label{eq:un}
u_n &\sim \text{Uniform}_{[0,1]},
&n&=1,\dots,N,
\\
\label{eq:sn}
s_n \mid B,h^{1:K} &\sim \text{Categorical}_{1:K}\!\left(\pi_{1:K}^G\right),
&n&=1,\dots,N.
\end{align}
In this formulation, the segment selection probabilities $\pi_{1:K}^G$ are determined by the geometry $G$ of the curve $\mathcal G$. Specifically, these are given by
\begin{align}
\label{eq:pi}
\pi_k^G
&=
\frac{
L_k^G
}{
\sum_{k'}L_{k'}^G
}
,&
k&=1,\dots,K,
\end{align}
where $L_k^G$ is the length of the $k$\textsuperscript{th} segment in the geometry $G$. This is computed by
\begin{align}
\label{eq:L}
L_k^G&=
\begin{cases}
\|h^k-h^{k+1}\|,  &\text{if  } k< B, \\
\|h^k-h^1\|, &\text{if  } k=B
,\\
0, &\text{if  } k> B,  \\
\end{cases}
&
k&=1,\dots,K,
\end{align}
where $\|\cdot\|$ denotes the Eucledian norm in $\mathbb{R}^d$.
As the following theorem shows, our hierarchical generation of the positions $r_{1:N}$ in \cref{eq:un,eq:sn,eq:pi,eq:L} is equivalent to \cref{eq:rn}.

\begin{theorem}
For a fixed $n=1,\dots,N$, let $r_n$ be a random position generated according to the sampling scheme in \cref{eq:un,eq:sn,eq:pi,eq:L}. Then \cref{eq:rn} follows.
\begin{proof}
Given a geometry $G$, the position $r_n$ specifies and is specified by a global coordinate $v_n$ that equals the arc-length from $h^1$ along the curve $\mathcal{G}$ encoded by $G$. Consequently, it suffices to show that $v_n$ is uniformly distributed over $[0,L^G]$ where $L^G=\sum_{k=1}^K L_k^G$ is the total arc-length of $\mathcal{G}$.

Conditional on $s_n$ and $u_n$, such global coordinate $r_n$ is obtained deterministically by
\begin{align*}
v_n|s_n,u_n
&\sim\delta_{C^G_{s_n-1}+u_nL_{s_n}^G},
\end{align*}
where the cumulative segment lengths are defined by
\begin{align*}
C^G_0&=0,
\\
C^G_k&=\sum_{k'=1}^K L^G_{k'},
&
k&=1,\dots,K.
\end{align*}
According to the law of total probability, we obtain the marginal
\begin{align*}
p(v_n)
&=
\sum_{s_n=1}^K\int_0^1du_n\,
p(v_n|s_n,u_n)p(s_n,u_n),
\end{align*}
and, because of independence in \cref{eq:sn,eq:un}, we obtain
\begin{align*}
p(v_n)
&=
\sum_{s_n=1}^K\int_0^1du_n\,
p(v_n|s_n,u_n)p(s_n)p(u_n)
\\
&=
\sum_{s_n=1}^K\int_0^1du_n\,
\delta_{C^G_{s_n-1}+u_nL_{s_n}^G}(v_n)\pi_{s_n}^G
\\
&=
\sum_{s_n=1}^K\int_0^1du_n\,
\frac{1}{L_{s_n}^G}\delta_{u_n}\left(\frac{v_n-C^G_{s_n-1}}{L_{s_n}^G}\right)\frac{L_{s_n}^G}{L^G}
\\
&=\frac{1}{L^G}
\sum_{s_n=1}^K\int_0^1du_n\,
\delta_{u_n}\left(\frac{v_n-C^G_{s_n-1}}{L_{s_n}^G}\right)
\\
&=\frac{1}{L^G}
\end{align*}
indicating that, indeed, $v_n$ is uniformly distributed over $[0,L^G]$ which completes our proof. Here, the last equality holds because, from the $K$ different summands $\int_0^1du_n\,
\delta_{u_n}\left(\frac{v_n-C^G_{s_n-1}}{L_{s_n}^G}\right)$, only one integrates over the singularity.
\end{proof}
\end{theorem}

\subsection{Bayesian considerations and non-parametric prior}

In order to allow for inference of the curve's latent geometry $G$, we adopt a non-parametric formulation in which the total number $K$  of model inducing points is infinite. To avoid overfitting, we assign a prior on the number of \emph{active} vertices of the form
\begin{align}
\label{eq:poiss}
B\sim\text{Poisson}(\gamma).
\end{align}
In this way, we ensure that although infinitely many model points are considered, those that are actually utilized in the geometry and eventually determine the curve $\mathcal{G}$, \ie the first $B$ of them as indicated by \cref{eq:L}, remain finite.

Finally, to each inducing point, we assign independent priors
\begin{align}
\label{eq:h_temp}
h^k &\sim \text{Normal}(\mu, \upsilon I),
&k&=1,\dots,K.
\end{align}

\begin{figure}[tbp]
\centering
\includegraphics[scale=0.7]{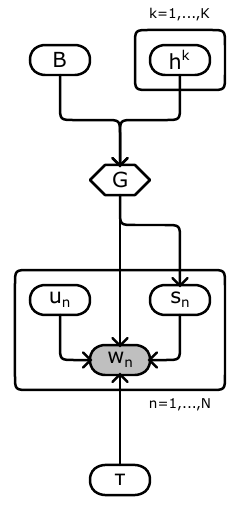}
\caption{Graphical representation of the statistical model in this study. Following the common convention \cite{presse2023data,bishop2006pattern }, random quantities are demonstrated with circles, deterministic quantities are shown with diamonds, and quantities with known values are shaded. In addition, arrows specify dependencies between the various quantities and plates indicate repetition over the specified index.}
\label{fig:graph}
\end{figure}

Our full statistical model combines the non-parametric prior on the inducing points, \cref{eq:poiss,eq:h_temp}, with the likelihood governing the observed point-cloud data, \cref{eq:norm_like}. Its complete set of equations is shown graphically in \cref{fig:graph} and summarized in
\begin{align}
\label{eq:full_1}
B &\sim \text{Poisson}(\gamma),
\\
\label{eq:full_2}
h^k &\sim \text{Normal}(\mu, \upsilon I),
& k&=1,\dots,K,
\\[1ex]
\label{eq:full_3}
u_n &\sim \text{Uniform}_{[0,1]},
& n&=1,\dots,N,
\\
\label{eq:full_4}
s_n \mid B,h^{1:K} &\sim \text{Categorical}_{1:K}\!\left(\pi_{1:K}^G\right),
& n&=1,\dots,N,
\\[1ex]
\label{eq:full_5}
w_n\mid r_n,\tau &\sim \text{Normal}\left(
r_n, I/\tau\right), & n&=1,\dots,N,
\\
\label{eq:full_6}
\tau &\sim \text{Gamma}(\phi, \tau_{\mathrm{ref}}/\phi),
\end{align}
where we have re-parameterized the noise variance $\sigma^2$ in terms of a precision parameter $\tau=1/\sigma^2$ and
adapted the standard prior to $\tau$. The hyperparameters $\gamma,\mu,\upsilon,\phi,\tau_{\mathrm{ref}}$ are fixed, and the selection probabilities $\pi^G_{1:K}$ are obtained according to \cref{eq:pi,eq:L}.

\subsection{Parametric model approximation}

The non-parametric formulation of our model \cite{muller2015bayesian, ghosal2017fundamentals} uses an infinite collection of inducing points to represent the curve $\mathcal{G}$ under reconstruction. Although modeling-wise this is a convenient representation, it is nevertheless computationally intractable, as non-parametric model formulations are associated with probability measures that do not attain probability functions \cite{ghosal2017fundamentals}. For this reason, we proceed with a finite approximation.

Specifically, we opt for a finite but exceedingly large number of candidate inducing points $K\gg 1$ and approximate \cref{eq:poiss} by
\begin{align}
\label{eq:B_B}
    B\sim \text{Binomial} \left(K,\gamma/K \right)
    .
\end{align}
Similarly to the exact non-parametric model, each of the $K$ inducing points in the approximate model is activated independently with probability $\gamma/K$. As a result, the expected number of active inducing points, in addition to being finite, remains equal to the expectation of the exact one $\mathbb{E}[B]=\gamma$. Furthermore, as we show below, our approximate finite model, mediated by \cref{eq:B_B}, at the limit $K\to\infty$, converges to the exact infinite model, mediated by \cref{eq:poiss}.

\begin{theorem}
\label{thm:binomial_poisson_approx}
At the limit $K\to\infty$, the variates generated according to \cref{eq:B_B} converge in distribution to the variates generated according to \cref{eq:poiss}. 
\begin{proof}
According to the binomial statistics of \cref{eq:B_B}, the probability mass function of $B$ is
\begin{align*}
p\left(B\right)
&= \frac{\gamma^B}{B!}\frac{K(K-1)\cdots (K-B+1)}{K^B}
\left(1-\frac{\gamma}{K}\right)^K
\left(1-\frac{\gamma}{K} \right)^{-B}.
\end{align*}
Due to the limits
\begin{align*}
\lim_{K\to\infty}\frac{K(K-1)\cdots(K-B+1)}{K^B} &=1, 
&
\lim_{K\to\infty}\left(1-\frac{\gamma}{K} \right)^K&=e^{-\gamma}, 
&
\lim_{K\to\infty}\left(1-\frac{\gamma}{K}\right)^{-B}&=1,
\end{align*}
we readily obtain the limiting probability mass
\begin{align*}
\lim_{K\to \infty} p(B)
= \frac{\gamma^B}{B!} e^{-\gamma},
\end{align*}
which is in agreement with the Poisson statistics of \cref{eq:poiss}. 
\end{proof}
\end{theorem}

\subsection{Markov chain Monte Carlo}

Following the parametric approximation of \cref{eq:poiss}, the posterior distribution of our model attains a probability function of the form
\begin{align*}
p\left(\tau, B, h^{1:K}, s_{1:N}, u_{1:N}\mid w_{1:N}\right)
.
\end{align*}
This can be used both for curve reconstruction and uncertainty quantification tasks, since it allows for the estimation of the geometry $G=\{B,h^{1:K}\}$ that uniquely determines a curve $\mathcal{G}$; nevertheless, it does not admit a convenient analytical expression that allows its direct characterization. To overcome this limitation, we employ Monte Carlo methods \cite{presse2023data,robert2004monte,gelman2013bayesian}, which enable posterior characterization through sampling simulations. 

\subsubsection{Sampler description}

To characterize the posterior $p\left(\tau, B, h^{1:K}, s_{1:N}, u_{1:N}\mid w_{1:N}\right)$, we develop a specialized Markov chain Monte Carlo (MCMC) sampler \cite{presse2023data} that iteratively updates our model variables. Our MCMC sampler targets the model posterior with an overall Gibbs sampling scheme \cite{robert2004monte,liu2001monte} consisting of the following two stages:
\begin{itemize}
\item Initialize $\tau,B,h^{1:K},u_{1:N}$ via random draws from their corresponding priors in \cref{eq:full_6,eq:B_B,eq:full_2,eq:full_3} and generate $s_{1:N}$ via \cref{eq:full_2}.
\item Iterate update steps on designated blocks of $\tau,B,h^{1:K},s_{1:N},u_{1:N}$ in a randomized order until a sufficient number of MCMC samples is generated.
\end{itemize}
However, due to strong dependencies between the posterior variables, univariate or small block updates in the iteration stage lead to slow mixing and poor exploration of the posterior space \cite{gelman2013bayesian,robert2004monte}, rendering naive MCMC impractical. To address this and achieve improved sampling efficiency, we design a sequence of increasingly refined samplers, starting from a basic implementation of blocked Gibbs updates and progressing to more advanced variants that incorporate structured model-specific updates that maintain marginalized out progressively larger variable blocks.

In detail, each variant of our MCMC sampler implements the iteration stage using different update mechanisms as listed below. 

\paragraph{Sampler 1} 
\begin{itemize}[label=-]
\item Update $\tau$ as in \emph{Update~A} below.
\item Jointly update $B, h^{1:K}, s_{1:N}, u_{1:N}$ as in \emph{Update~B} below.
\end{itemize}

\paragraph{Sampler 2}
\begin{itemize}[label=-]
\item Update $\tau$ as in \emph{Update~A} below.
\item Jointly update $B, h^{1:K}, s_{1:N}, u_{1:N}$ as in \emph{Update~B} below.
\item Jointly update $B, s_{1:N}, u_{1:N}$ as in \emph{Update~C} below. 
\end{itemize}

\paragraph{Sampler 3}
\begin{itemize}[label=-]
\item Jointly update $\tau,h^{1:K}$ as in \emph{Update~A*} below. 
\item Jointly update $B, h^{1:K}, s_{1:N}, u_{1:N}$ as in \emph{Update~B} below.
\item Jointly update $B,h^{1:K},s_{1:N},u_{1:N}$ as in \emph{Update~B*} below.
\item Jointly update $B, s_{1:N}, u_{1:N}$ as in \emph{Update~C} below.
\item Jointly update $h^{1:K},s_{1:N},u_{1:N}$ as in \emph{Update~D} below.
\end{itemize}

\subsubsection{Update mechanisms}
\label{sec:updates}

We describe our individual update mechanisms below. These consist of a combination of ancestral sampling (AS) \cite{liu2001monte}, direct sampling (DS) \cite{robert2004monte}, elliptical slice sampling (ESS) \cite{nishihara2014parallel}, and custom Metropolis-Hastings (MH) methods \cite{gelman2013bayesian,robert2004monte}. A detailed description, along with derivations and implementation specifics, is deferred to the \textsc{Appendix}, see \cref{sec:sub_SI,sec:sub_SI_0}.

\paragraph{Update A} 
The target is $p\left(\tau|B,h^{1:K},s_{1:N},u_{1:N},w_{1:N}\right)$. This target is sampled via DS (see \cref{sec:sub_SI_1}).

\paragraph{Update B} 
The target is $p\left(
B,h^{1:K},s_{1:N},u_{1:N}|\tau,w_{1:N}
\right)$. This is sampled via AS according to the factorization
\begin{align*}
p\left(
B,h^{1:K},s_{1:N},u_{1:N}|\tau,w_{1:N}
\right)
&=
p\left(
u_{1:N}|B,h^{1:K},s_{1:N},\tau,w_{1:N}
\right)
\\
&\times p\left(
s_{1:N}|B,h^{1:K},\tau,w_{1:N}
\right)
\\
&\times p\left(
B,h^{1:K}|\tau,w_{1:N}
\right).
\end{align*}
The first and second factors are sampled via DS (see \cref{sec:sub_SI_2,sec:sub_SI_3}), and the third factor is sampled via ESS and DS after completion with auxiliary variables (see \cref{sec:sub_SI_4}).

\paragraph{Update C} 
The target is $p\left(
B,s_{1:N},u_{1:N}|h^{1:K},\tau,w_{1:N}
\right)$. This is sampled via AS according to the factorization
\begin{align*}
p\left(
B,s_{1:N},u_{1:N}|h^{1:K},\tau,w_{1:N}
\right)
&=
p\left(
u_{1:N}|B,h^{1:K},s_{1:N},\tau,w_{1:N}
\right)
\\
&\times p\left(
s_{1:N}|B,h^{1:K},\tau,w_{1:N}
\right)
\\
&\times p\left(
B|\tau,h^{1:K},w_{1:N}
\right).
\end{align*}
The first and second factors are sampled as in \emph{Update~B} (see \cref{sec:sub_SI_2,sec:sub_SI_3}), and the third factor is sampled via DS (see \cref{sec:sub_SI_5}).

\paragraph{Update D} 
The target is $p\left(
h^{1:K},s_{1:N},u_{1:N}|B,\tau,w_{1:N}
\right)$. This is sampled via AS according to the factorization
\begin{align*}
p\left(
h^{1:K},s_{1:N},u_{1:N}|B,\tau,w_{1:N}
\right)
&=
p\left(
u_{1:N}|B,h^{1:K},s_{1:N},\tau,w_{1:N}
\right)
\\
&\times p\left(
s_{1:N}|B,h^{1:K},\tau,w_{1:N}
\right)
\\
&\times p\left(
h^{1:B}|B,h^{B+1:K},\tau,w_{1:N}
\right)\\
& \times p\left(h^{B+1:K}|B,\tau,w_{1:N}\right).
\end{align*}
The first and second factors are sampled as in \emph{Update~B} (see \cref{sec:sub_SI_2,sec:sub_SI_3}), the third factor is sampled via a custom MH that randomly permute, reflect, and reconnect the geometry's vertices (see \cref{sec:sub_SI_6}), and the fourth factor is sampled via DS (see \cref{sec:sub_SI_7}).

\paragraph{Update A*} 
The target is $p\left(\tau,h^{1:K}|B,s_{1:N},u_{1:N},w_{1:N}\right)$. This is sampled via AS according to the factorization
\begin{align*}
p\left(\tau,h^{1:K}|B,s_{1:N},u_{1:N},w_{1:N}\right)
&= p\left(\tau|h^{1:B},s_{1:N},u_{1:N},w_{1:N}\right)
\\
&\times 
p\left(h^{1:B}|B,s_{1:N},u_{1:N},w_{1:N}\right)\\
&\times p\left( h^{B+1:K}|B,s_{1:N},u_{1:N},w_{1:N}\right).
\end{align*}
The first factor is sampled as in \emph{Update~A} (see \cref{sec:sub_SI_1}), the second factor is sampled via ESS (see \cref{sec:sub_SI_8}), and the third factor is sampled via DS (see \cref{sec:sub_SI_9}).

\paragraph{Update B*}
The target is the same as in \emph{Update~B} and is sampled via AS according to the same factorization. Sampling of the first and second factors is unchanged (see \cref{sec:sub_SI_2,sec:sub_SI_3}); however, sampling of the third factor is performed via a custom MH that randomly modifies the geometry by adding or removing vertices after completion with auxiliary variables (see \cref{sec:sub_SI_10}).

\subsubsection{Overview}

From our three MCMC variants, Sampler~1 is mainly based on ESS in \emph{Update~B} that instantly changes only the position of the active inducing points $h^{1:B}$ in a given geometry $G$. Although theoretically able to recover the correct geometry in the long run, in practice convergence is slow due to the local nature of the ESS which often results in trapping in local posterior modes corresponding to curves only partially matching the point-cloud under analysis.

To this end, Sampler~2 introduces additional joint moves in \emph{Update~C} that involve inducing points and latent variables to facilitate instant changes in the positioning of active inducing points $h^{1:B}$ and also their total number $B$. These improve mixing and generally lead to better exploration of the posterior space compared to Sampler~1. However, trapping in local modes persists as reordering of the active vertices needed to escape partially matching curves, even with this sampler, requires multiple trials.

Finally, Sampler~3 introduces specialized MH moves in \emph{Updates~B* and~D} that instantly change the positioning, ordering, and total number of active inducing points $h^{1:B}$. These result in changes that allow for transitioning between local posterior modes and substantially improved mixing. Although these moves affect the geometry $G$ globally, they actually do not drastically modify the reconstructed curve $\mathcal{G}$, which can alternate between partially matching configurations while maintaining nearly constant data likelihoods. For this reason, the corresponding MH proposals, despite their global nature, are accepted with high probability. As a result, Sampler~3 exhibits faster convergence at relatively low cost and no tendency to get trapped in local modes such as Samplers~1 and 2.

\subsection{Data acquisition}

All point-clouds used in the numerical illustrations presented in \textsc{Results} contain 2D localizations and were obtained from three sources: (i) synthetic data generated from our proposed model, (ii) curated point-cloud datasets available in the literature, and (iii) real-world data from LiDAR land surveying studies.

Synthetic data were generated by simulating \cref{eq:norm_like,eq:rn} under prescribed geometries $G=\{B,h^{1:K}\}$ for $K\gg1$ as well as prescribed noise $\tau$ and size $N$. The prescribed geometries were retained and used only post hoc as references (\ie ground-truth) for comparison against the MCMC reconstructions.

Curated point-clouds were obtained from benchmark studies \cite{ohrhallinger2018stretchdenoise,ohrhallinger2019fitconnect,lee2000curve}. These include the \textsc{Butterfly example} from Refs.~\cite{ohrhallinger2019fitconnect,ohrhallinger2018stretchdenoise}, the \textsc{Sawtooth example} from Ref.~\cite{ohrhallinger2018stretchdenoise}, and the \textsc{Bottle example} from Ref.~\cite{lee2000curve}. 

Real-world point clouds were obtained from the \textsc{Mapping Rock Glaciers (2025)} survey \cite{ruef2026rockglaciers} and the \textsc{U.S.~Geological 3D elevation (2016)} program \cite{usgs3dep2016}. The dataset from the former consists of boundary points mapping the derived outlines of rock glaciers extracted from airborne LiDAR point returns in the San Juan Mountains, Colorado, USA. The dataset from the latter consists of downsampled classified (land, water, and vegetation) points delineating the shoreline of McBee Island extracted from airborne LiDAR measurements in the Holston River, TN, USA. 



\section{Results}

Having described our reconstruction model and its computational implementation in \textsc{Methods}, we now show selected results that demonstrate its characteristics and performance. We begin by verifying that our proposed methodology generates valid curve reconstructions by comparing directly against the ground-truth using synthetic datasets of varying complexity. Using synthetic data, we also conduct ladders of simulations to demonstrate that our framework meaningfully quantifies and propagates uncertainty from point-clouds to reconstructions and compare the performance of our three MCMC samplers, highlighting their relative efficiency and reconstruction accuracy. Finally, we compare our results with standardized benchmark datasets from \cite{ohrhallinger2019fitconnect,ohrhallinger2018stretchdenoise,lee2000curve} as well as with field-derived LiDAR data from  \cite{ruef2026rockglaciers,usgs3dep2016}.

\subsection{Illustration with synthetic data}

To validate our reconstruction framework, we tested it on two synthetically generated datasets that can be seen in \cref{fig:lissajous,fig:octagon}. Additional results are provided in the \textsc{Appendix}, see \cref{app:add_res}. To quantify the accuracy of the reconstructed curves, for all cases, we compute the Hausdorff distance \cite{chazal2006sampling} between the ground-truth and each MCMC posterior sample. This distance is given by
\begin{align*}
d\left(\mathcal{G}_*,\mathcal{G}\right)
=\max\left\{
\max_{g_*\in\mathcal{G}_*}\min_{g\in\mathcal{G}}\|g_*-g\|
,
\max_{g\in\mathcal{G}}\min_{g_*\in\mathcal{G}_*}\|g-g_*\|
\right\}
\end{align*}
where $\mathcal{G}_*$ denotes the ground-truth curve and $\mathcal{G}$ denotes the curve reconstructed via the geometry $G=\{B,h^{1:K}\}$ of a sample in the generated MCMC chain. As can be seen, $d\left(\mathcal{G}_*,\mathcal{G}\right)$ provides a global measure of the discrepancy between the two curves \cite{chazal2006sampling}, with $d\left(\mathcal{G}_*,\mathcal{G}\right)=0$ indicating a perfect match and $d\left(\mathcal{G}_*,\mathcal{G}\right)>0$ quantifying the degree of mismatch.

In \cref{fig:lissajous} we evaluate our framework on a challenging dataset generated from a Lissajous curve \cite{lawrence2013catalog}, defined by
\begin{align*}
\mathcal{G}=\left\{\left(\sin(3t),\sin(2t)\right)\right\}_{t\in[0,2\pi)}\subset\mathbb{R}^2,
\end{align*}
which contains numerous self-intersections and intricate geometric features. Such a non-convex geometry provides a demanding benchmark for assessing the sampler's ability to recover complex curve structures from noisy point-cloud observations. \Cref{fig:lissajous} summarizes posterior sampled curves and the maximum a posteriori (MAP) reconstruction. The close agreement between the MAP estimate and the ground-truth, together with the concentration of the posterior samples around the ground-truth, demonstrates that the proposed Bayesian framework accurately reconstructs highly complex curves. The accompanying posterior distribution of the number of active vertices $B$ indicates that the model is able to identify and adjust sufficiently many vertices to match the geometry, and the low Hausdorff distance further indicates that the model reliably matches the ground-truth globally.

\begin{figure}[tbp]
\includegraphics[scale=0.7]{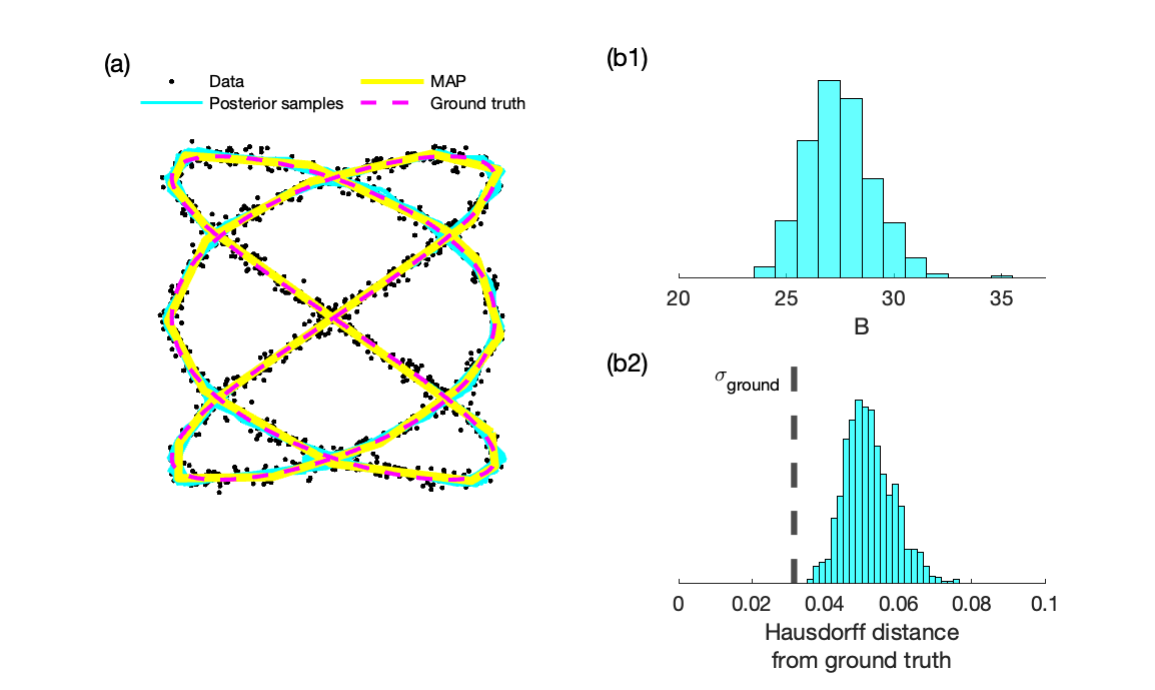}
\centering
\caption{Synthetic data analysis of a Lissajous curve.
Reconstruction is applied on a noisy point-cloud of size $N=1500$ and noise $\tau=1000$ (\ie $\sigma\approx0.032$). 
A total of 3,000 MCMC iterations are performed using Sampler~3, with the first 50\% discarded as burn-in and the remaining samples thinned at a ratio of 1:3 for visualization purposes.
The panel on the left shows the point-cloud and reconstructed curves. The panels to the right summarize key posterior results.} 
\label{fig:lissajous}
\end{figure}

In \cref{fig:octagon} we evaluate the proposed framework on a synthetic dataset generated from a regular octagon defined by the inducing points
\begin{align*}
\mathcal{H}=
\left\{
\left(\cos\left(k\pi/4\right),
\sin\left(k\pi/4\right)\right)
\right\}_{k=0,1,\ldots,7}\subset \mathbb{R}^{2}.
\end{align*}
Compared with the Lissajous example in \cref{fig:lissajous}, this geometry provides a simpler benchmark with well-defined edges and vertices, allowing us to verify that the sampler accurately recovers polygonal structures despite discontinuities at the corners. \Cref{fig:octagon} summarizes the posterior sampled curves and the maximum a posteriori (MAP) reconstruction. Again, the close agreement between the MAP estimate and the ground-truth, together with the concentration of the posterior samples around the ground-truth, demonstrates that the proposed Bayesian framework accurately reconstructs polygonal curves. The posterior distribution of the number of active vertices $B$ is centered near the ground-truth value, while the Hausdorff distance remains consistently small, indicating excellent agreement between the inferred and ground-truth curves.

\begin{figure}[tbp]
\includegraphics[scale=0.7]{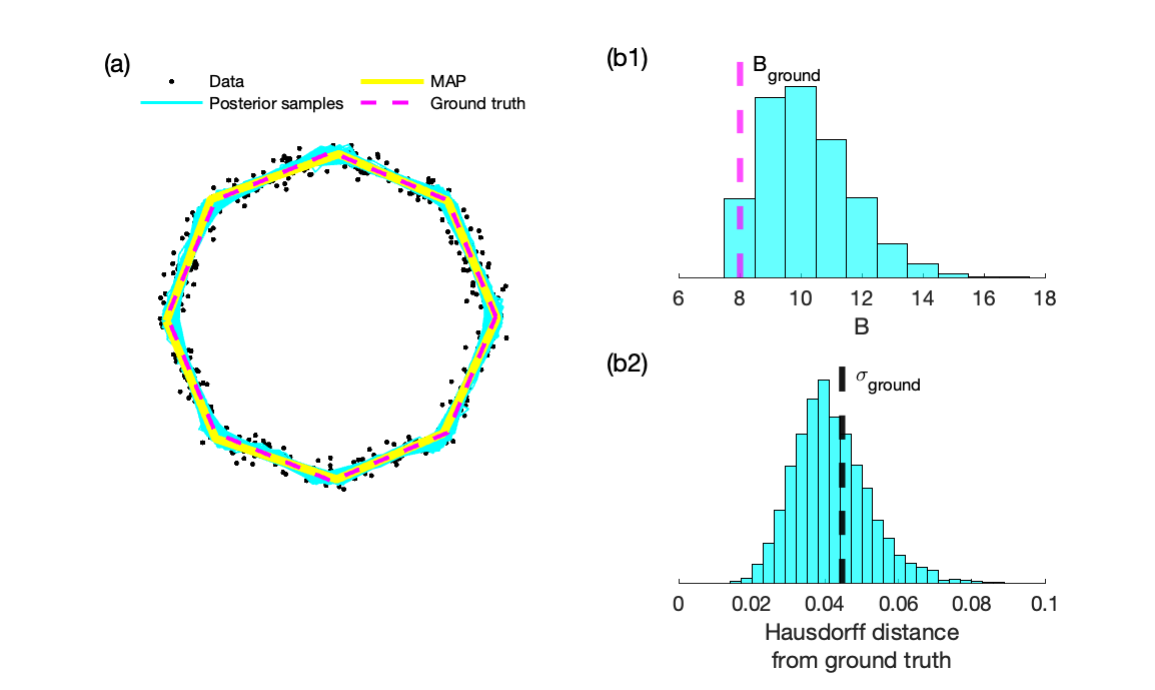}
\centering
\caption{Synthetic data analysis of an octagon curve. Reconstruction is performed on a noisy point-cloud of size $N=12000$ and noise $\tau=500$ (\ie $\sigma\approx0.045$). A total of 12,000 MCMC iterations are performed using Sampler~3, with the first 25\% discarded as burn-in and the remaining samples thinned at a ratio of 1:3 for illustration purposes. The panel on the left shows the point-cloud and reconstructed curves. The panels to the right summarize key posterior results.
}
\label{fig:octagon}
\end{figure}

\subsection{Uncertainty propagation}

\begin{figure}[tbp]
\includegraphics[scale=0.7]{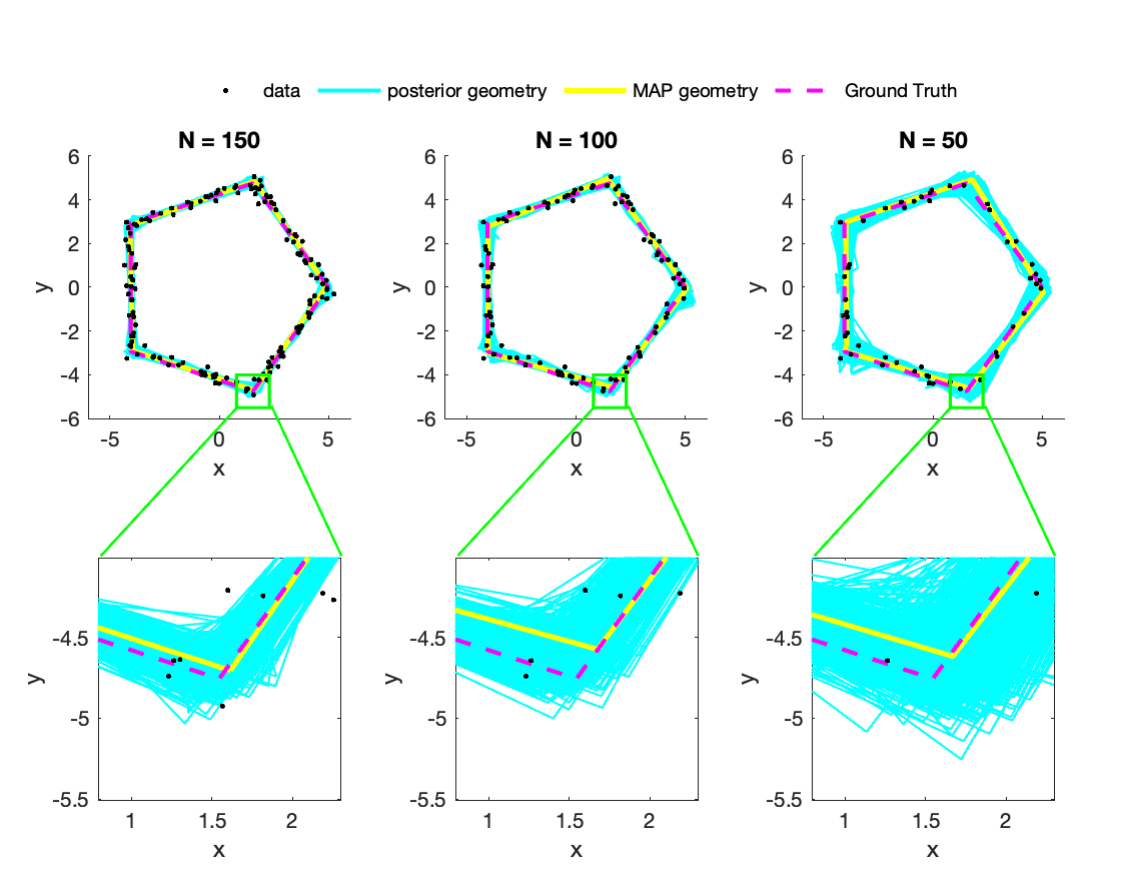}
\centering
\caption{Effects of point-cloud size. Reconstruction is performed on three point-clouds with sizes $N=150,100,50$, left to right, and fixed noise $\tau=30$ (\ie $\sigma\approx0.18$). For each experiment, $3{,}000$ MCMC iterations are performed using Sampler~3, with the first 30\% discarded as burn-in and the remaining samples thinned at a ratio of 1:2 for illustration purposes. The top panels show the point-clouds used and reconstructed curves. The square marks the same neighborhood of the lower-right vertex in each panel, and the bottom panels show a magnified view of this region using identical axis limits. The enlarged panels highlight the increase in posterior uncertainty as point-cloud size decreases.}
\label{fig:sa_N}
\end{figure}

\begin{figure}[tbp]
\includegraphics[scale=0.7]{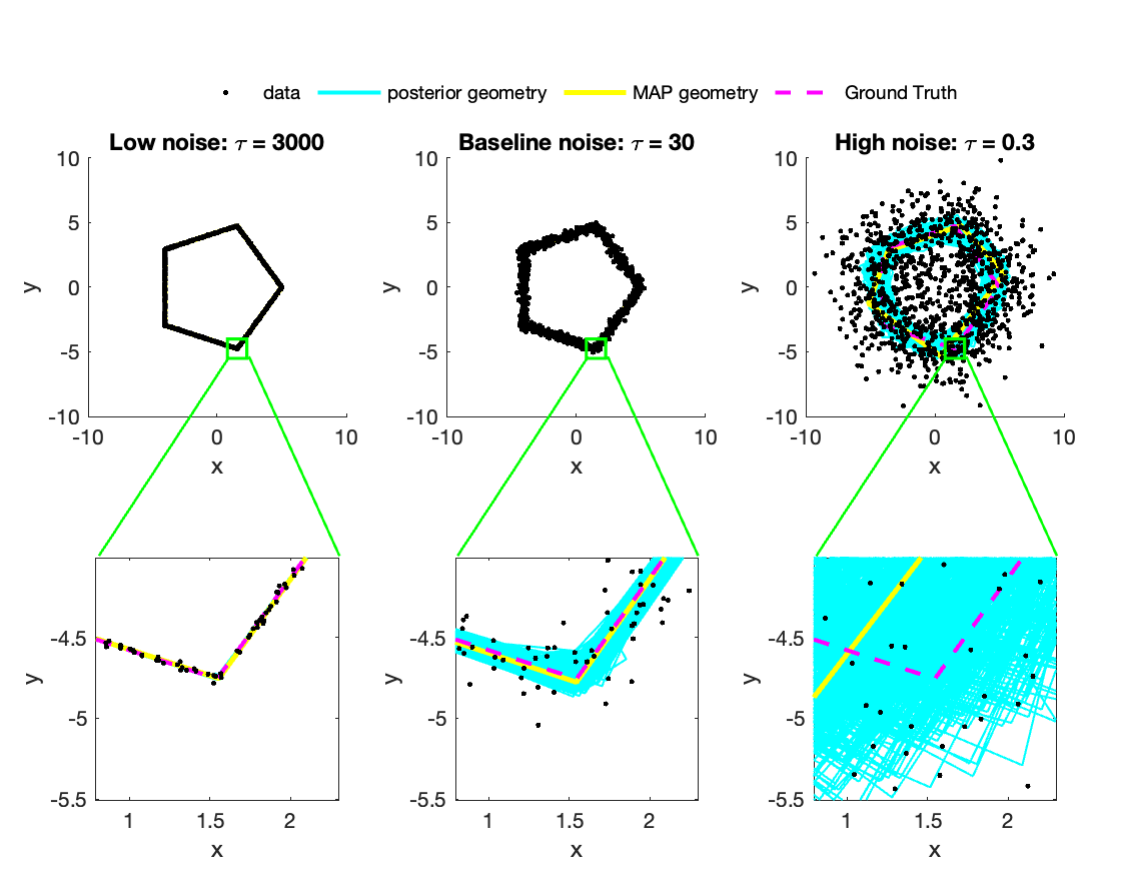}
\centering
\caption{Effects of localization noise. Reconstruction is performed on point-clouds of size $N=1000$ and noise levels $\tau=3000,30,0.3$ (\ie $\sigma\approx0.018,0.18,1.8$), left to right. For each experiment, $3{,}000$ MCMC iterations are performed using Sampler~3, with the first 30\% discarded as burn-in and the remaining samples thinned at a ratio of 1:3 for visualization purposes. 
The top panels show the point-clouds used and reconstructed curves. The square marks the same neighborhood of the lower-right vertex in each panel, and the bottom panels show a magnified view of this region using identical axis limits. The enlarged panels highlight the increase in posterior uncertainty as noise increases.}
\label{fig:sa_t}
\end{figure}

Since our framework relies on a Bayesian model, one of its main characteristics is that it can propagate uncertainty from a given dataset to the resulting reconstructions. In the context of curve reconstruction, the uncertainty of the data is influenced by the total number of localizations in the point-cloud $N$, that affects the degree of missing information, and the noise level $\tau$, that affects the corruption caused by localization noise. In \cref{fig:sa_N,fig:sa_t} we provide sensitivity analyzes with respect to both factors. For these tests, we evaluate the proposed framework on a synthetic datasets generated from a regular pentagon centered at the origin and defined by the vertices
\begin{align*}
\mathcal{H}=
\left\{
\left(
5\cos\left(2k\pi/5\right),
5\sin\left(2k\pi/5\right)
\right)
\right\}_{k=0,1,\ldots,4}
\subset \mathbb{R}^{2}.
\end{align*}

In \cref{fig:sa_N}, we start with a point-cloud of reference size $N=150$ and progressively reduce it to $N=100$ and $N=50$ with a fixed noise $\tau =30$ or $\sigma =.18$. As size decreases, the obtained posterior widens, as expected, indicating increased uncertainty over the resulting reconstructions.

In \cref{fig:sa_t}, we start with a point-cloud corrupted with a reference noise of $\tau=3000$ that corresponds to a standard deviation of $\sigma\approx0.018$ and progressively reduce it to $\tau=30$ and $\tau=0.3$ that correspond to standard deviations of $\sigma\approx0.18$ and $\sigma\approx1.8$, respectively. Again, as noise increases, the obtained posterior widens, as expected, indicating increased uncertainty over the resulting reconstructions.

\subsection{Comparison of sampling schemes}

In \textsc{Methods} we presented three variants of our MCMC algorithm for the evaluation of our posterior reconstructions that differ in complexity and sophistication. To compare the effectiveness of the three variants, we apply each sampler to the same synthetic point-cloud and evaluate its ability to recover the underlying ground-truth curve.

\begin{figure}[tbp]
\includegraphics[scale=1.7]{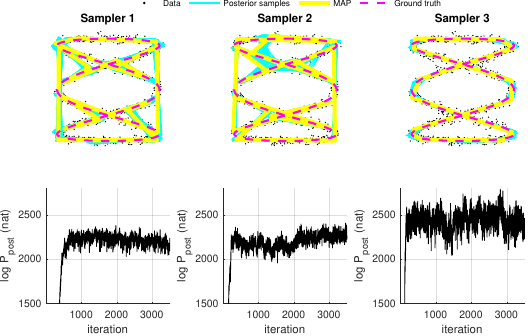}
\centering
\caption{Comparison of MCMC samplers on the same synthetic point-cloud. The top panels show the point-clouds used and reconstructed curves. The bottom panels show traces of the corresponding log-posterior. From the three samplers, only Sampler~3 recovers the ground-truth; while, Samplers 1 and 2 remains trapped in local maxima corresponding to only partially matching configurations.}
\label{fig:compare}
\end{figure}

\Cref{fig:compare} shows that, while all three samplers target the same posterior distribution, their mixing behavior differs substantially. Thanks to its specialized MH updates that re-position, re-order, and re-size a geometry's active vertices, Sampler~3 explores the posterior space more effectively reaching higher posterior values and avoiding the partially matching configurations that attract Samplers~1 and 2. As a result, Sampler~3 produces the most robust and accurate inference.

More concretely, the three variants lead to Hausdorff distances between the ground-truth and MAP reconstructions of $0.312$, $0.318$, and $0.075$ for Samplers~1, 2, and 3, respectively. Once again, this indicates that only reconstructions from Sampler~3 closely match the ground-truth.

\subsection{Benchmark with standardized point-clouds}

\begin{figure}[tbp]
\includegraphics[scale=0.7]{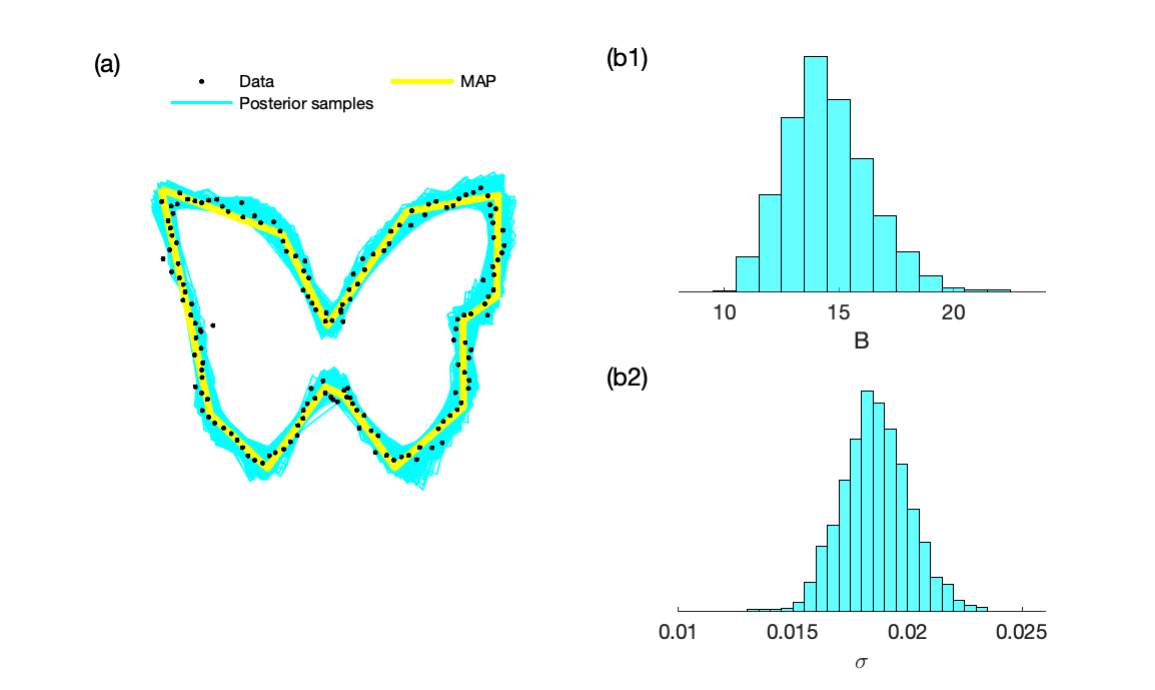}
\centering
\caption{Reconstruction of the \textsc{Butterfly example}.
A total of 5,000 MCMC iterations are performed using Sampler~3, with the first 50\% discarded as burn-in and the remaining samples thinned at a ratio of 1:2 to improve visual clarity.
The panel on the left shows the point-cloud and reconstructed curves. The panels to the right summarize key posterior results.
}
\label{fig:butterfly}
\end{figure}

To highlight our method's reconstruction characteristics, we consider standardized point-clouds from the \textsc{FitConnect} and \textsc{StretchDenoise} studies \cite{ohrhallinger2019fitconnect,ohrhallinger2018stretchdenoise} as well as Ref.~\cite{lee2000curve}. We include examples from these three studies because they represent state-of-the-art reconstruction methods and have all been evaluated previously on the same or similar standardized point-clouds. \Cref{fig:butterfly} shows reconstructions obtained from the \textsc{Butterfly example}, which has previously been used as a benchmarking test for \textsc{FitConnect} and \textsc{StretchDenoise} and additional results on the \textsc{Sawtooth} and \textsc{Bottle examples} are provided in the \textsc{Appendix}, see \cref{app:add_res}.

Of the three methods considered, \textsc{FitConnect} relies on a geometric approach designed to connect noisy 2D point samples by fitting parametric arcs to local circular point neighborhoods. In this way, \textsc{FitConnect} can recover the connectivity (\ie sequential ordering) of unorganized points, analogous to the sequential ordering of the inducing points in our framework that form the vertices of the curve under reconstruction. Following a similar approach, \textsc{StretchDenoise} extends \textsc{FitConnect} by using its connectivity as an intermediate stage, on which it applies separate parametric adjustments to optimize the vertices' positions within approximately derived local bounds. In contrast, Ref.~\cite{lee2000curve} presents a deterministic curve-reconstruction method based on moving least squares and Euclidean minimum spanning trees. This method uses iterative refinement of the original point-cloud to progressively thin its localizations until a smooth non-intersecting curve emerges.

As explained above, all three methods provide deterministic curve reconstructions that best match the characteristics of point-estimates \cite{gelman2013bayesian} that most closely resemble our MAP curves. However, rather than returning a single reconstructed curve, our framework generates an ensemble of posterior samples over admissible curve reconstructions. We use the posterior to obtain a representative reconstruction (\eg a MAP estimate), while also quantifying the uncertainty around this curve. In contrast, uncertainty quantification for the resulting reconstruction is absent in the other methods. In addition, our framework explicitly accounts for localization noise in the original point-cloud which is estimated along the other quantities of interest, and its uncertainty is propagated to the resulting reconstructions in contrast to the mentioned methods where noise is incorporated as a generic optimization device to enable local or global curve refinements. Finally, our framework proceeds in a single stage where connectivity (\ie sequential ordering), positioning, and sizing (\ie total number) of the active inducing points forming the vertices of the curve under reconstruction are interrelated, allowing information from one to influence the others rather than utilizing multiple separate stages that prohibit such influences.

\subsection{Applications on field-data}

To demonstrate the application of our framework to real-world data, we consider the analysis of two datasets obtained through airborne LiDAR acquisition techniques, as commonly found in environmental and geospatial studies. First, we use a point-cloud from the \textsc{Mapping Rock Glaciers (2025)} survey \cite{ruef2026rockglaciers}. Our selected dataset consist of boundary points that represent the outlines of mapped rock glaciers in the San Juan Mountains, CO. Second, we use a point-cloud from the \textsc{U.S.~Geological Survey 3D Elevation} program \cite{usgs3dep2016}. Our selected dataset represents the outline of McBee Island in the Holston river, TN.

\begin{figure}[tbp]
\includegraphics[scale=.7]{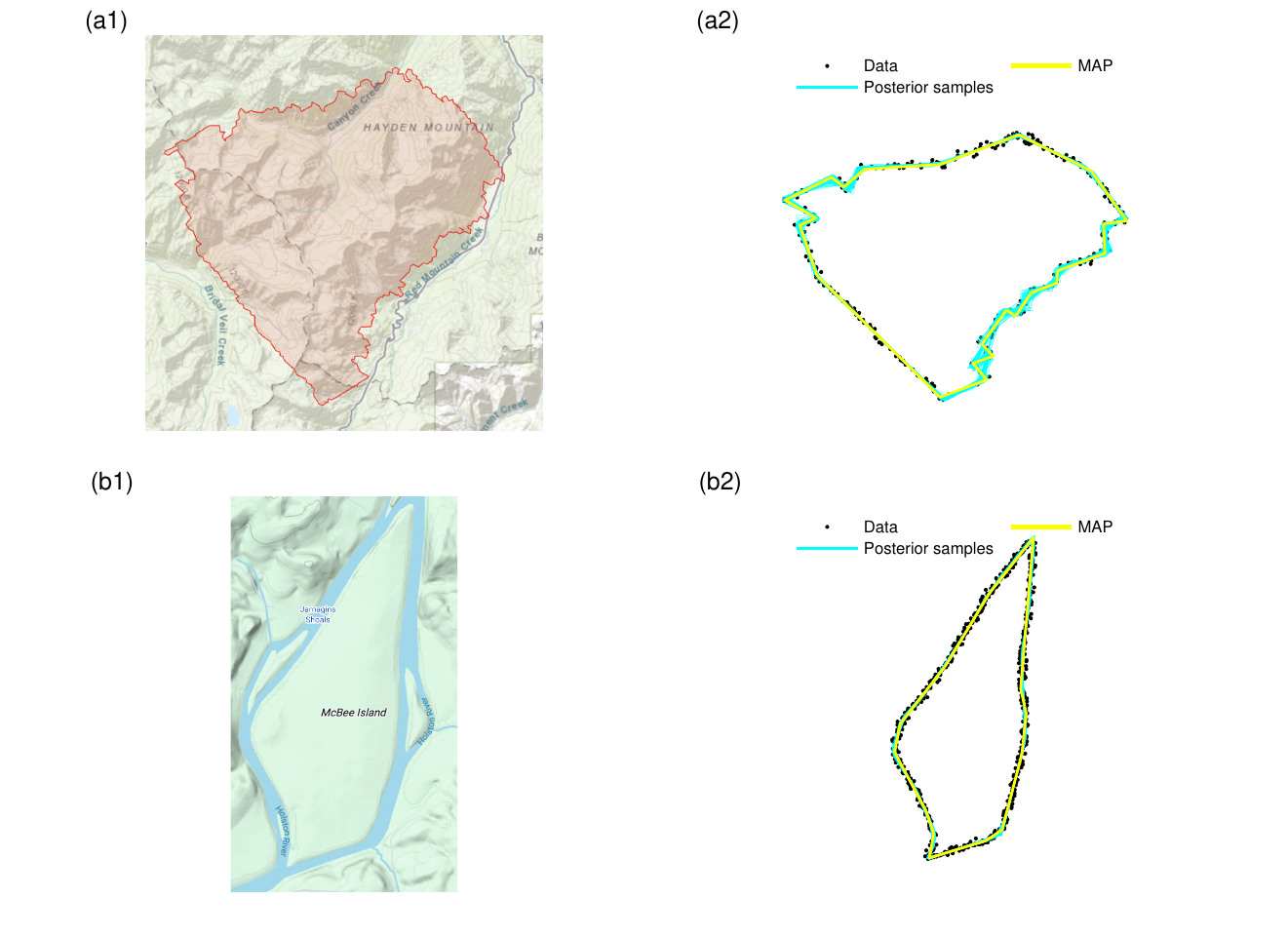}
\centering
\caption{Reconstruction of field LiDAR point-clouds imaging rock glaciers in the San Juan Mountains, CO (top panels) and McBee Island in the Holston River, TN (bottom panels). In each case, a total of 3,000 MCMC iterations are performed using Sampler~3, with the first 50\% discarded as burn-in and the remaining samples thinned at a ratio of 1:8 for visualization purposes. The left panels show reference outlines of the imaged structures and the right panels show the point-cloud used and reconstructed curves.}
\label{fig:lidar_CO_TN}
\end{figure}

As seen in \cref{fig:lidar_CO_TN}, starting from each point-cloud, our Bayesian reconstruction framework recovers the underlying geometric structure (rock glacier in (a1)-(a2) and McBee Island in (b1)-(b2)) while accounting for data uncertainty. Both reconstructions closely follow the observed point-clouds and produce smooth, coherent boundary estimates despite the sparse and non-uniform sampling of the original datasets. In addition, unlike deterministic reconstructions, our framework also provides posterior uncertainty quantification. This is particularly pronounced in undersampled regions, for example, upper-left and lower-right sides in (a2), or regions of high curvature, for example, upper and lower corners in (b2).

Overall, these two LiDAR case studies show that our Bayesian framework reliably reconstructs geospatial structures while explicitly quantifying reconstruction uncertainty. The posterior uncertainty naturally increases in undersampled and high-curvature regions, providing a readily interpretable measure of the regions where the inferred geometry is most uncertain.

\section{Discussion}

In this study, we presented a novel comprehensive, fully Bayesian framework for point-cloud data analysis aimed specifically at tasks involving curve reconstruction under uncertainty. Our proposed methodology integrates a principled probabilistic model for the curve's latent geometry with an explicit representation of localization noise and data corruption, together with an array of computational strategies that make posterior evaluation efficient and tractable for practical applications. As a result, our approach supports both high-fidelity reconstruction and uncertainty quantification from incomplete, sparse, or degraded point-cloud observations and can become an indispensable component in AI/ML, data analysis, or image processing workflows operating on curated or raw localization measurements.

Relative to existing point-cloud reconstruction pipelines that are often dominated by deterministic fitting\cite{ohrhallinger2019fitconnect,han2025implicit}, heuristic regularization\cite{ohrhallinger2018stretchdenoise,peng2022method,ritter2021robust}, or purely geometric criteria\cite{lee2000curve,gu2007ricci,liu2022operator}, our formulation provides a unified statistical treatment of data, missing information, and reconstruction ambiguity. In particular, rather than producing a single curve, \ie a point estimate, our approach follows Bayesian principles and yields a posterior distribution over admissible reconstructions, enabling the derivation of credible intervals, probabilistic error bars, and uncertainty-aware downstream decisions\cite{gelman2013bayesian}. This contrasts with many widespread methods in which uncertainty is ignored\cite{ritter2021robust} or added post hoc, and robustness to data corruption can depend sensitively on hyperparameter fine-tuning and other user choices\cite{ohrhallinger2019fitconnect,ohrhallinger2018stretchdenoise,ohrhallinger20212d}.

Nevertheless, despite the advantages of our methodology, two limitations remain important. First, while feasible, the computational cost can still be substantial because our computational strategy relies on repetitive Markov chain Monte Carlo (MCMC) sampling, which may become a bottleneck for large point-clouds or time-critical applications. Second, our current computational design is tailored to one-dimensional geometries, these include curves embedded in 2D, 3D, or higher-dimensional Euclidean scenes, and does not directly extend to the reconstruction of surfaces or volumetric structures, where the dimensionality of the latent object and the complexity of the associated likelihood integrals increase markedly.

Such limitations could be addressed in several directions. For instance, the computational burden of MCMC could be reduced through high-performance computing (especially in updates A*, B*, C, and D of \cref{sec:updates}), including parallelization techniques and distributed processing, that exploit the natural structure of our likelihood---which maintains independence over data points and segments---to speed up the completion of each MCMC iteration. In addition, extending our methodology beyond curves may be possible by introducing numerical approximations for the higher-dimensional integrals mediating the factorization of the model posterior (see \textsc{Appendix}, \cref{sec:sub_SI}) that arise in surface/volume settings, or by developing surrogate models (\eg emulators or reduced-order representations) that preserve essential geometric and uncertainty features while making inference scalable. However, fully resolving these challenges requires methodological and implementation advances beyond the scope of a single proof-of-concept study and will be the focus of future research.


\appendix

\section{Main posterior factorization and definitions}
\label{sec:sub_SI}

Sampling of most targets described in \cref{sec:sub_SI_0} relies on the following factorization of the model's conditional posterior:
\begin{align*}
p\left(w_{1:N},u_{1:N},s_{1:N},h^{1:K},B|\tau\right)
&=
p\left(w_{1:N},u_{1:N},s_{1:N},h^{1:K}|B,\tau\right)
p\left(B|\tau\right)
\\
&=
p\left(w_{1:N},u_{1:N},s_{1:N},h^{1:K}|B,\tau\right)
p\left(B\right)
\\
&=
p\left(w_{1:N},u_{1:N},s_{1:N}|B,h^{1:K},\tau\right)
p\left(h^{1:K}|B,\tau\right)
p\left(B\right)
\\
&=
p\left(w_{1:N},u_{1:N},s_{1:N}|B,h^{1:K},\tau\right)
p\left(h^{1:K}\right)
p\left(B\right)
\\
&=
p\left(w_{1:N},u_{1:N},s_{1:N}|B,h^{1:K},\tau\right)
\left[\prod_kp\left(h^k\right)\right]
p\left(B\right)
\\
&=
\left[
\prod_n
p\left(w_n,u_n,s_n|B,h^{1:K},\tau\right)
\right]
\left[\prod_kp\left(h^k\right)\right]
p\left(B\right)
\\
&=
\left[
\prod_n
p\left(w_n,u_n,s_n|G,\tau\right)
\right]
\left[\prod_kp\left(h^k\right)\right]
p\left(B\right)
\\
&=
\left[
\prod_n
\mathcal{F}^{G,\tau}_n(u_n,s_n)
\right]
\left[\prod_k\text{Normal}\left(h^k;\mu,\upsilon I\right)\right]
\text{Binomial}\left(B;K,\gamma/K\right)
.
\end{align*}
The function $\mathcal{F}^{G,\tau}_n(u_n,s_n)$ is defined and factorized as follows:
\begin{align*}
\mathcal{F}^{G,\tau}_n(u_n,s_n)
&=
p\left(w_n,u_n,s_n|G,\tau\right)
\\
&=
p\left(w_n|u_n,s_n,G,\tau\right)
p\left(u_n|s_n,G,\tau\right)
p\left(s_n|G,\tau\right)
\\
&=
p\left(w_n|u_n,s_n,G,\tau\right)
p\left(u_n\right)
p\left(s_n|G\right)
\\
&=
p\left(s_n|G\right)
p\left(w_n|u_n,s_n,G,\tau\right)
p\left(u_n\right)
\\
&=
\text{Categorical}_{1:K}\left(s_n;\pi^G_{1:K}\right)
\text{Normal}_2\left(w_n;r^G_{s_n,u_n},I_2/\tau\right)
\text{Uniform}_{[0,1]}\left(u_n\right)
\\
&=
\frac{\tau}{2\sqrt{2\pi}}
\,
F^G_n
\,
\text{Categorical}_{1:K}\left(s_n;f_n^{1:K}\right)
\,
\text{Normal}_{[0,1]}\left(u_n;M_n^{s_n},\left(S^{s_n}\right)^2\right).
\end{align*}
Here, the constants, which depend implicitly on the geometry $G=\{B,h^{1:K}\}$, are given by:
\begin{align*}
q_n^s&=h^s-w_n,
\\
p^s&=h^s-h^{s+1},
\\
S^s&=\frac{1}{\|p^s\|\sqrt{\tau}},
\\
M_n^s&=\frac{(p^s)^tq^s_n}{\|p^s\|^2},
\\
\Omega_n^s&=\pi^G_sS^s
\exp\left(\frac{(M_n^s/S^s)^2-\tau\|q_n^s\|^2}{2}\right)
\left[
\erf\left(\frac{1}{\sqrt{2}}\frac{1-M_n^s}{S^s}\right)
+
\erf\left(\frac{1}{\sqrt{2}}\frac{M_n^s}{S^s}\right)
\right],
\\
F^G_n&=\sum_{s'}\Omega_n^{s'},
\\
f^s_n&=\frac{\Omega_n^s}{F^G_n}.
\end{align*}
The computation of $\Omega_n^s$ requires the computation of the sum
\begin{align*}
\erf\left(\frac{1}{\sqrt{2}}\frac{1-M_n^s}{S^s}\right)
+
\erf\left(\frac{1}{\sqrt{2}}\frac{M_n^s}{S^s}\right)
=
\erfc\left(\frac{1}{\sqrt{2}}\frac{\left|M_n^s-\frac{1}{2}\right|-\frac{1}{2}}{S^s}\right)
-
\erfc\left(\frac{1}{\sqrt{2}}\frac{\left|M_n^s-\frac{1}{2}\right|+\frac{1}{2}}{S^s}\right)
\end{align*}
which is prone to catastrophic cancellation and numerical underflow. For this reason, we implement the right-hand side instead, which is numerically robust.

\section{Sampling of individual MCMC targets}
\label{sec:sub_SI_0}

\subsection{Target in Updates~A and~A*}
\label{sec:sub_SI_1}

The target is $p\left(\tau|B,h^{1:K},s_{1:N},u_{1:N},w_{1:N}\right)$ and is derived as follows:
\begin{align*}
p\left(
\tau|B,h^{1:K},s_{1:N},u_{1:N},w_{1:N}
\right)&=p\left(\tau|h^{1:B},s_{1:N},u_{1:N},w_{1:N}\right)\\
&\propto p\left(w_{1:N}|h^{1:B},s_{1:N},u_{1:N},\tau\right)p\left(\tau|h^{1:B},s_{1:N},u_{1:N}\right)\\
&=p\left(w_{1:N}|h^{1:B},s_{1:N},u_{1:N},\tau\right)p\left(\tau\right)\\
    &=\left(\prod_{n=1}^{N} p\left( w_{n}|h^{1:B},s_{n},u_{n},\tau \right) \right) p(\tau)\\
    &=\left(\prod_{n=1}^{N} \text{Normal}\left(w_{n};r_{s_{n},u_{n}}^{G},I/\tau \right) \right) \text{Gamma}\left(\tau; \phi, \frac{\tau_{\text{ref}}}{\phi} \right)\\
    & \propto \text{Gamma}\left(\tau; \phi+\frac{d}{2} N, \frac{1}{\frac{\phi}{\tau_{\rm ref}}+\frac{1}{2}\sum_{n=1}^{N}\left\|w_{n}-r_{s_{n},u_{n}}^{G}\right\|^{2}} \right)
    \\
& =\text{Gamma}\left(\tau; \phi',\beta'\right)
\end{align*}
This target is sampled via DS. The constants are defined by
\begin{align*}
\phi'&=\phi+\frac{d}{2} N,
\\
\beta'&=\frac{1}{\frac{\phi}{\tau_{\rm ref}}+\frac{1}{2}\sum_{n=1}^{N}\left\|w_{n}-r_{s_{n},u_{n}}^{G}\right\|^{2}} .
\end{align*}

\subsection{Target in Updates~B,~C,~D, and~B*}
\label{sec:sub_SI_2}

The target is $p\left(
u_{1:N}|B,h^{1:K},s_{1:N},\tau,w_{1:N}
\right)$ and, due to the factorization in \cref{sec:sub_SI}, is derived as follows:
\begin{align*}
p\left(u_{1:N}|B,h^{1:K},s_{1:N},\tau,w_{1:N}\right)
&\propto
p\left(w_{1:N},u_{1:N},s_{1:N},h^{1:K},B|\tau\right)
\\
&\propto
\prod_n
\text{Normal}_{[0,1]}\left(u_n;M_n^{s_n},\left(S^{s_n}\right)^2\right)
\end{align*}
This target is sampled via DS \cite{botev2017normal}. Here, $\text{Normal}_{[0,1]}$ denotes the truncated normal distribution.

\subsection{Target in Updates~B,~C,~D, and~B*}
\label{sec:sub_SI_3}

The target is $p\left(
s_{1:N}|B,h^{1:K},\tau,w_{1:N}
\right)$ and, due to the factorization in \cref{sec:sub_SI}, is derived as follows:
\begin{align*}
p\left(s_{1:N}|B,h^{1:K},\tau,w_{1:N}\right)
&\propto
p\left(w_{1:N},s_{1:N},h^{1:K},B|\tau\right)
\\
&=\int du_{1:N}\,
p\left(w_{1:N},u_{1:N},s_{1:N},h^{1:K},B|\tau\right)
\\
&\propto
\prod_n
\text{Categorical}_{1:K}\left(s_n;f_n^{1:K}\right)
\end{align*}
This target is sampled via DS.

\subsection{Target in Update~B}
\label{sec:sub_SI_4}

The target is $p\left(
B,h^{1:K}|\tau,w_{1:N}
\right)$ and is first completed with a discrete auxiliary variable $b$ as follows:
\begin{align*}
p\left(
B,h^{1:K}|\tau,w_{1:N}
\right)
&=
\sum_bp\left(
b,B,h^{1:K}|\tau,w_{1:N}
\right).
\end{align*}
The auxiliary random variable is obtained so as to satisfy:
\begin{align*}
b|B\sim\text{Categorical}_{B-1,B,B+1}\left(\frac{1}{3},\frac{1}{3}, \frac{1}{3} \right)
.
\end{align*}
Subsequently, the completed target $p\left(b,B,h^{1:K}|\tau,w_{1:N}\right)$ is sampled with two Gibbs stages:

\textbullet~Update $b$ from the target $p(b|B,h^{1:K},\tau,w_{1:N})$. This target is derived as follows
\begin{align*}
p(b|B,h^{1:K},\tau,w_{1:N})&=p(b|B)
\\
&=\text{Categorical}_{B-1,B,B+1}\left(b;\frac{1}{3},\frac{1}{3}, \frac{1}{3} \right)
\end{align*}
and sampled with DS.

\textbullet~Update $B,h^{1:K}$ jointly from the target $p(B,h^{1:K}|b,\tau,w_{1:N})$. This target is factorized as follows:
\begin{align*}
p\left(B,h^{1:K}|b,\tau,w_{1:N}\right)
&=
p\left(B|h^{1:K},b,\tau,w_{1:N} \right)
p\left(h^{1:K}|b,\tau,w_{1:N} \right)
\end{align*}
and sampled with AS. Specifically, due to the factorization in \cref{sec:sub_SI}, the last factor is derived as follows:
\begin{align*}
p\left(h^{1:K}|b,\tau,w_{1:N} \right)
&=\sum_{B=0}^{K} p\left(h^{1:K},B |b,\tau,w_{1:N}\right)
\\ 
&\propto \sum_{B=0}^{K} p\left(b,h^{1:K},B|\tau,w_{1:N} \right)
\\ 
&=\sum_{B=0}^{K} p\left(b|h^{1:K},B,\tau,w_{1:N} \right)p\left(h^{1:K},B|\tau,w_{1:N} \right)
\\
&=\sum_{B=0}^{K}p\left(b|B \right) p\left(h^{1:K},B|\tau,w_{1:N} \right)
\\
&=\sum_{B=b-1}^{b+1}\frac{1}{3}p\left(h^{1:K},B|\tau,w_{1:N} \right)
\\
&\propto
\sum_{B=b-1}^{b+1}p\left(w_{1:N},h^{1:K},B|\tau\right)
\\
&=
\sum_{B=b-1}^{b+1}\int du_{1:N}\sum_{s_{1:N}}p\left(w_{1:N},u_{1:N},s_{1:N},h^{1:K},B|\tau\right)
\\
&\propto
\sum_{B=b-1}^{b+1}\left[\prod_{n}F_{n}^{G} \right]\left[\prod_k\text{Normal}\left( h^{k};\mu,\upsilon I\right) \right] \text{Binomial}\left(B;K,\gamma/K\right)
\\
&=
\left[\prod_k\text{Normal}\left( h^k;\mu,\upsilon I\right) \right]
\sum_{B=b-1}^{b+1} \left[\prod_{n}F_{n}^{G} \right] \text{Binomial}\left(B;K,\gamma/K\right) \end{align*}
and, because of the normal factors, is sampled with ESS. Finally, the first factor, due to the factorization in \cref{sec:sub_SI}, is derived as follows:
\begin{align*}
p\left(B|h^{1:K},b,\tau,w_{1:N} \right)
&\propto
p\left(b|w_{1:N},h^{1:K},B,\tau \right)
p\left(B|w_{1:N},h^{1:K},\tau \right)
\\
&=
p\left(b|B\right)
p\left(B|w_{1:N},h^{1:K},\tau \right)
\\
&\propto
p\left(b|B \right) p\left(w_{1:N},h^{1:K},B|\tau \right)
\\
&=
p\left(b|B \right)\int du_{1:N}\sum_{s_{1:N}} p\left(w_{1:N},u_{1:N},s_{1:N},B,h^{1:K}|\tau \right)
\\
&\propto
\begin{cases}
\left[\prod_{n} F_{n}^{G}\right]\text{Binomial}\left(B;K,\gamma/K \right) &\text{if  } B\in \{b-1,b,b+1 \}
\\
0 &\text{otherwise}
\end{cases}
\\ 
&\propto
\text{Categorical}_{0:K}\left(B;\omega_{0:K}\right)
\end{align*}
and sampled with DS. The constants $\omega_{0:K}$ are given by
\begin{align*}
\omega_B&=
\begin{cases}
\frac{\prod_nF_n^G\text{Binomial}\left(B;K,\gamma/K \right)}{\sum_{B'=b-1}^{b+1} \prod_nF_n^{G'}\text{Binomial}\left(B';K,\gamma/K \right)} &\text{if  } B\in \{b-1,b,b+1 \}
\\
0 &\text{otherwise}
\end{cases}
,
&
B&=0,\dots,K.
\end{align*}

\subsection{Target in Update~C}
\label{sec:sub_SI_5}

The target is $p\left(
B|\tau,h^{1:K},w_{1:N}
\right)$ and, due to the factorization in \cref{sec:sub_SI}, is derived as follows:
\begin{align*}
p\left(B|h^{1:K},\tau,w_{1:N}\right)
&\propto
p\left( w_{1:N},h^{1:K},B|\tau\right)
\\
&=
\int du_{1:N}\sum_{s_{1:N}}
p\left( w_{1:N},u_{1:N},s_{1:N},h^{1:K},B|\tau\right)
\\
&\propto\left[\prod_nF_n^G\right]\text{Binomial}\left(B;K,\gamma/K \right)
\\ 
&\propto
\text{Categorical}_{0:K}\left(B;\omega_{0:K}'\right)
\end{align*}
This target is sampled via DS. The constants $\omega_{0:K}'$ are similar to \cref{sec:sub_SI_4} and are given by
\begin{align*}
\omega_B'
&=
\frac{\prod_nF_n^G\text{Binomial}\left(B;K,\gamma/K \right)}{\sum_{B'} \prod_nF_n^{G'}\text{Binomial}\left(B';K,\gamma/K \right)},
&
B&=0,\dots,K.
\end{align*}

\subsection{Target in Update~D}
\label{sec:sub_SI_6}

The target is
$p\left(h^{1:B}\middle|B,h^{B+1:K},\tau,w_{1:N}\right)$
and, due to the factorization in \cref{sec:sub_SI}, is derived as follows:
\begin{align*}
p\left(h^{1:B}\middle|B,h^{B+1:K},\tau,w_{1:N}\right)
&\propto
p\left(w_{1:N},h^{1:K},B\middle|\tau\right)
\\
&=
\int du_{1:N}\sum_{s_{1:N}}
p\left(
w_{1:N},u_{1:N},s_{1:N},h^{1:K},B
\middle|\tau
\right)
\\
&\propto
\left[\prod_n F_n^G\right]
\left[
\prod_k
\text{Normal}
\left(h^k;\mu,\upsilon I_2\right)
\right].
\end{align*}
This target is sampled via a custom MH. Our MH mixes randomly, with equal probability, two proposal distributions:
\begin{align*}
h^{1:B}\prop|h^{1:B}\old&\sim\frac{1}{2}\mathbb{Q}_1(h^{1:B}\old)+\frac{1}{2}\mathbb{Q}_2(h^{1:B}\old)
\end{align*}
The first proposal, $\mathbb{Q}_1$, applies a circular shift to the vertices $h^{1:B}\old$ followed by a reconnection. The circular shift is performed in either the negative or positive direction with equal probability. First, the circular shift is applied for a number of steps chosen from $\{1,\dots,B-1\}$ uniformly at random. Then, the reconnection generates $h^{1:B}\prop$ by choosing two distinct positions $k',k''$ uniformly at random from $\{2,\dots,B-1\}$ and reversing the order of the vertices from $\min(k',k'')$ through $\max(k',k'')$, including both endpoints.

The second proposal, $\mathbb{Q}_2$, starts by reversing the vertices in $h^{1:B}\old$. With equal probability, either an inducing point based reflection or a segment based reflection is used, and the anchor index is chosen uniformly from $1$ to $B$.  Followed by the same reconnection operation as in proposal $\mathbb{Q}_1$.

Both $\mathbb{Q}_1$ and $\mathbb{Q}_2$ are symmetric MCMC proposals since they implement isometries between $h^{1:B}\old$ and $h^{1:B}\prop$, and for this reason the acceptance of our MH sampler reduces to that of the Metropolis sampler \cite{robert2004monte,presse2023data}. In addition, since the actual placement of the vertices does not differ between $h^{1:B}\old$ and $h^{1:B}\prop$, but only their arrangement, the second factor in the target above drops out.

\subsection{Target in Update~D}
\label{sec:sub_SI_7}

The target is $p\left(h^{B+1:K}|B,\tau,w_{1:N}\right)$ and is derived as follows:
\begin{align*}
p\left(h^{B+1:K}|B,\tau,w_{1:N}\right)
&=
p\left(h^{B+1:K}\right)
\\
&=
\prod_{k=B+1}^K
p\left(h^k\right)
\\
&=
\prod_{k=B+1}^K
\text{Normal}\left(h^k;\mu,\upsilon I\right)
\end{align*}
This target is sampled via DS.

\subsection{Target in Update~A*}
\label{sec:sub_SI_8}

The target is $p\left(h^{1:B}|B,s_{1:N},u_{1:N},w_{1:N}\right)$ and is derived as follows:
\begin{align*}
p\left(h^{1:B}|B,s_{1:N},u_{1:N},w_{1:N}\right) &=\int_{0}^{\infty} p\left(\tau,h^{1:B}|B,s_{1:N},u_{1:N},w_{1:N}\right) d\tau
\\
&\propto\int_{0}^{\infty} p\left(B,s_{1:N},u_{1:N},w_{1:N}|\tau,h^{1:B}\right)p\left(\tau,h^{1:B}\right)d\tau
\\ 
&=\int_{0}^{\infty} p\left(w_{1:N}|\tau,h^{1:B},B,s_{1:N},u_{1:N}\right)  p\left(B,s_{1:N},u_{1:N}|\tau, h^{1:B}\right) p\left(\tau\right) p\left(h^{1:B}\right) d\tau
\\
&\propto \int_{0}^{\infty} p\left(w_{1:N}|\tau,h^{1:B},B,s_{1:N},u_{1:N}\right) p\left(s_{1:N}|B,h^{1:B}\right)
p\left(\tau\right)
p\left(h^{1:B}\right) d\tau \\
&\propto p\left(h^{1:B}\right) p\left(s_{1:N}|B,h^{1:B}\right)
\int_{0}^{\infty} p\left(w_{1:N}|\tau,h^{1:B},B,s_{1:N},u_{1:N}\right) p\left(\tau \right) d\tau
\\
&\propto p\left(h^{1:B}\right) p\left(s_{1:N}|B,h^{1:B}\right)\left(\beta'\right)^{\phi'}
\\
&=\left[\prod_{k=1}^B\text{Normal}\left(h^k;\mu,\upsilon I\right)\right]
\left[\prod_n\pi^G_{s_n}\right]
\left(\beta'\right)^{\phi'}
\end{align*}
This target is sampled via ESS. The weights $\pi^G_{1:B}$ are given in \cref{eq:pi}, the constants $\phi'$ and $\beta'$ are the same as in \cref{sec:sub_SI_1}, and the integral is given by 
\begin{align*}
\int_{0}^{\infty} p\left(w_{1:N}|\tau,h^{1:B},B,s_{1:N},u_{1:N}\right) p\left(\tau\right) d\tau
&=
\int_{0}^{\infty} \prod_{n=1}^{N} \text{Normal}\left( w_{n}; r_{s_{n},u_{n}}^{G},\frac{I}{\tau}\right) \text{Gamma} \left(\tau; \phi, \frac{\tau_{\rm ref}}{\phi} \right) d\tau
\\
&\propto
\int_{0}^{\infty} \tau^{\phi'-1}\exp \left( -\frac{\tau}{\beta'}\right) d\tau
\\ 
&=\Gamma(\phi') \left(\beta'\right)^{\phi'} \int_{0}^{\infty}\text{Gamma}(\tau;\phi' ,\beta') d\tau
\\ 
&\propto \left(\beta'\right)^{\phi'}.
\end{align*}

\subsection{Target in Update~A*}
\label{sec:sub_SI_9}

The target is $p\left(h^{B+1:K}|B,s_{1:N},u_{1:N},w_{1:N}\right)$ and is derived as follows:
\begin{align*}
p\left(h^{B+1:K}|B,s_{1:N},u_{1:N},w_{1:N}\right)
&=
p\left(h^{B+1:K}\right)
\\
&=
\prod_{k=B+1}^K
p\left(h^k\right)
\\
&=
\prod_{k=B+1}^K
\text{Normal}\left(h^k;\mu,\upsilon I\right)
\end{align*}
This target is sampled via DS. The implementation is similar to \cref{sec:sub_SI_7}.

\subsection{Target in Update~B*}
\label{sec:sub_SI_10}

The target is $p\left(
B,h^{1:K}|\tau,w_{1:N}
\right)$ and is completed first with auxiliary variables $g_1,g_2,\sigma,\gamma$ as follows:
\begin{align*}
p\left(
B,h^{1:K}|\tau,w_{1:N}
\right)
&=
\int dg_1\int dg_2\sum_\sigma\int d\gamma\,
p\left(g_1,g_2,\sigma,\gamma,
B,h^{1:K}|\tau,w_{1:N}
\right).
\end{align*}
The auxiliary random variables are obtained by:
\begin{align*}
g_1&\sim\text{Gamma}\left(1,1\right),
\\
g_2&\sim\text{Gamma}\left(1,1\right),
\\
\sigma&\sim\text{Categorical}_{+1,-1}\left(\frac{1}{2},\frac{1}{2}\right),
\\
\gamma&\sim\frac{1}{2}\text{Beta}^{+1}\left(A_1,A_2\right)+\frac{1}{2}\text{Beta}^{-1}\left(A_1,A_2\right).
\end{align*}
Here, $\text{Beta}^{+1}$ and $\text{Beta}^{-1}$ denote the beta and reciprocal beta distributions \cite{presse2023data}, and $A_1,A_2$ are tunable parameters which we set to default values of $A_{1}=1$ and $A_{2}=125$. Subsequently, the completed target $p\left(g_1,g_2,\sigma,\gamma,B,h^{1:K}|\tau,w_{1:N}\right)$ is sampled with two Gibbs stages:

\textbullet~Update $g_1,g_2,\sigma,\gamma$ from the target $p(g_1,g_2,\sigma,\gamma|B,h^{1:K},\tau,w_{1:N})$. This target is derived as follows
\begin{align*}
p(g_1,g_2,\sigma,\gamma|B,h^{1:K},\tau,w_{1:N})
&=
p(g_1,g_2,\sigma,\gamma)
\\
&=
p(g_1)p(g_2)p(\sigma)p(\gamma)
\\
&=
\text{Gamma}\left(g_1;1,1\right)
\text{Gamma}\left(g_2;1,1\right)
\text{Categorical}_{+1,-1}\left(\sigma;\frac{1}{2},\frac{1}{2}\right)
\\
&\times\left[\frac{1}{2}\text{Beta}^{+1}\left(\gamma;A_1,A_2\right)+\frac{1}{2}\text{Beta}^{-1}\left(\gamma;A_1,A_2\right)
\right]
\end{align*}
and sampled with DS.

\textbullet~Update $B,h^{1:K}$ jointly from the target $p(B,h^{1:K}|g_1,g_2,\sigma,\gamma,\tau,w_{1:N})$. Due to the factorization in \cref{sec:sub_SI}, this target is derived as follows:
\begin{align*}
p(B,h^{1:K}|g_1,g_2,\sigma,\gamma,\tau,w_{1:N})
&=
p(B,h^{1:K}|\tau,w_{1:N})
\\
&\propto
p(w_{1:N},B,h^{1:K}|\tau)
\\
&=
\sum_{s_{1:N}}
\int du_{1:N}\, 
p(w_{1:N},u_{1:N},s_{1:N},B,h^{1:K}|\tau)
\\
&\propto
\left[\prod_{n}F_{n}^{G} \right]\left[\prod_k\text{Normal}\left( h^{k};\mu,\upsilon I\right) \right] \text{Binomial}\left(B;K,\gamma/K\right)
\end{align*}
and sampled with a custom MH. Specifically, our MH uses proposals from a distributional family
\begin{align*}
B\prop,h^{1:K}\prop|B\old,h^{1:K}\old
&\sim
Q^{g_1,g_2,\sigma,\gamma}\left(B\old,h^{1:K}\old\right)
\end{align*}
which allows factorization of the form:
\begin{align*}
Q^{g_1,g_2,\sigma,\gamma}\left(B\prop,h^{1:K}\prop|B\old,h^{1:K}\old\right)
&=
Q^{g_1,g_2,\sigma,\gamma}\left(h^{1:K}\prop|B\old,h^{1:K}\old,B\prop\right)
\\
&\times
Q^{g_1,g_2,\sigma,\gamma}\left(B\prop|B\old,h^{1:K}\old\right)
.
\end{align*}
To achieve add/remove proposals with high acceptance rate, we specialize our sampler by imposing further assumption on each factor.

\begin{itemize}
\item \textbf{Assumption 1:} For the second factor of the proposal, we assume:
\begin{align*}
Q^{g_1,g_2,\sigma,\gamma}\left(B\prop|B\old,h^{1:K}\old\right)
=
Q\left(B\prop|B\old\right)
.
\end{align*}
That is, the proposals of $B$ are independent of $h^{1:K}$ and also independent of the auxiliary parameters $g_1,g_2,\sigma,\gamma$. In addition, we assume that:
\begin{align*}
Q\left(B\prop|B\old\right)
=
\text{Categorical}_{B\old-1,B\old+1}\left(B\prop;\frac{1}{2},\frac{1}{2}\right)
.
\end{align*}

\item \textbf{Assumption 2i:} For the first factor of the proposal, we assume:
\begin{align*}
Q^{g_1,g_2,\sigma,\gamma}\left(h^{1:K}\prop|B\old,h^{1:K}\old,B\prop\right)
&=
\mathbb{F}^{g_1,g_2,\sigma,\gamma}_{B\old\to B\prop}\left(h^{B\old}\prop|h^{1:K}\old\right)
\\
&\times
\mathbb{G}^{g_1,g_2,\sigma,\gamma}_{B\old\to B\prop}\left(h^{B\prop}\prop|h^{1:K}\old\right)
\\
&\times
\prod_{k\neq B\old,B\prop}
\delta_{h^k\old}\left(h^k\prop\right)
\end{align*}
for appropriate distributions $\mathbb{F}^{g_1,g_2,\sigma,\gamma}_{\beta\to\beta'}\left(h^{\beta}\prop|h^{1:K}\old\right)$ and $\mathbb{G}^{g_1,g_2,\sigma,\gamma}_{\beta\to\beta'}\left(h^{\beta'}\prop|h^{1:K}\old\right)$ that are specified in the following.

\item \textbf{Assumption 2ii:} We assume only distributions $\mathbb{F}^{g_1,g_2,\sigma,\gamma}_{\beta\to\beta'}\left(h\prop|h^{1:K}\old\right)$ that are related to the distributions $\mathbb{G}^{g_1,g_2,\sigma,\gamma}_{\beta\to\beta'}\left(h\prop|h^{1:K}\old\right)$ like this:
\begin{align*}
\mathbb{F}^{g_1,g_2,\sigma,\gamma}_{\beta\to\beta'}\left(h\prop|h^{1:K}\old\right)
=
\mathbb{G}^{g_1,g_2,\sigma,1/\gamma}_{\beta'\to\beta}\left(h\prop|h^{1:K}\old\right)
\end{align*}
Note that ``add'' proposals in $\mathbb{G}$ correspond to ``remove'' proposals in $\mathbb{F}$, and that ``dines-in'' in $\mathbb{G}$ corresponds to ``dives-out'' in $\mathbb{F}$.

\item \textbf{Assumption 2iii:} We assume only deterministic distributions $\mathbb{G}^{g_1,g_2,\sigma,\gamma}_{\beta\to\beta'}\left(h\prop|h^{1:K}\old\right)$ that are defined like this:
\begin{align*}
\mathbb{G}^{g_1,g_2,\sigma,\gamma}_{\beta\to\beta'}\left(h\prop|h^{1:K}\old\right)
=
\begin{cases}
\delta_{R^{g_1,g_2,\sigma,\gamma}\left(h^1\old,h^\beta\old,h^{\beta'}\old\right)}\left(h\prop\right)
&\beta'=\beta+1
\\[2ex]
\delta_{h\old^{\beta'}}\left(h\prop\right)&\beta'=\beta-1
\end{cases}
\end{align*}
for an appropriate function $R^{g_1,g_2,\sigma,\gamma}(h',h'',h''')$ that is specified below.

\item \textbf{Assumption 2iv:} Finally, we assume a function $R^{g_1,g_2,\sigma,\gamma}(h',h'',h''')$ that is defined like this:
\begin{align*}
R^{g_1,g_2,\sigma,\gamma}(h',h'',h''')
=
\frac{g_1 h'+g_2 h''}{g_1+g_2}
+\sigma\gamma
\left(
h'''-\frac{g_1 h'+g_2 h''}{g_1+g_2}
\right)
\end{align*}
This function attracts or expels $h'''$ according to the dive determined by the $\gamma$ and the sign $\sigma$ around an anchor determined by $h',h''$ and the weights $g_1,g_2$.

\end{itemize}

\begin{figure}[ht]
\centering
\input{proposal_diagram.tex}
\caption{Add/remove proposal mechanism used in \emph{Update~B*}. The left and right panels illustrate the cases $B_{\rm prop}=B_{\rm old}+1$ (add) and $B_{\rm prop}=B_{\rm old}-1$ (remove), respectively. The complementary proposals $\mathbb{G}$ and $\mathbb{F}$ applied in the two cases determine the positioning of vertex $h^B$, while the remaining inducing points remain unchanged.}
\label{fig:app_add_remove}
\end{figure}
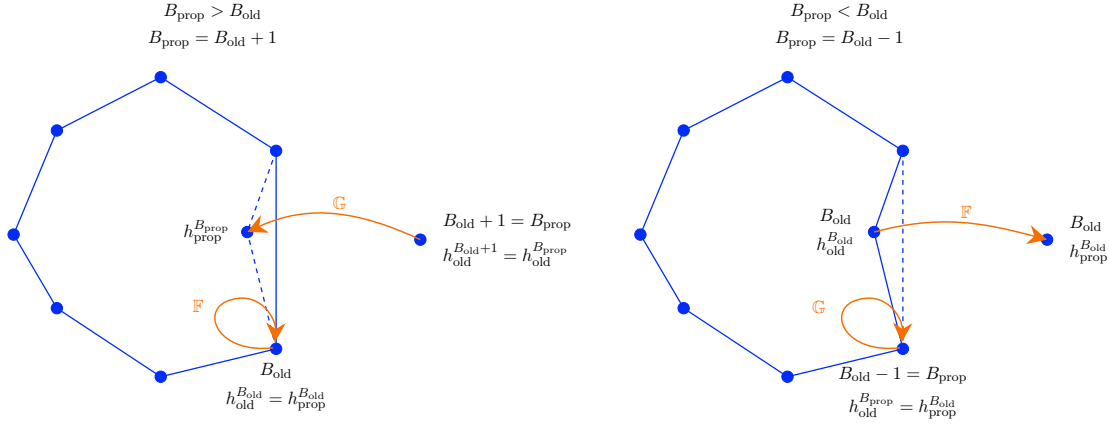

Our entire add/remove proposal mechanism is shown schematically on \cref{fig:app_add_remove}. The resulting MH sampler involves an acceptance ratio of the form:
\begin{align*}
A
&=
\underbrace{
\frac{
p\left(B\prop,h^{1:K}\prop|g_1,g_2,\sigma,\gamma,\tau,w_{1:N}\right)
}{
p\left(B\old,h^{1:K}\old|g_1,g_2,\sigma,\gamma,\tau,w_{1:N}\right)
}
}_{A_p}
\times
\underbrace{
\frac{
Q^{g_1,g_2,\sigma,\gamma}\left(B\old,h^{1:K}\old|B\prop,h^{1:K}\prop\right)
}{
Q^{g_1,g_2,\sigma,\gamma}\left(B\prop,h^{1:K}\prop|B\old,h^{1:K}\old\right)
}
}_{A_Q}
\end{align*}
which is readily computed. Specifically, $A_p$ is obtained by the ratio of the targets 
\begin{align*}
A_p&=
\dfrac{\left[\prod_{n}F_{n}^{G} \right]_{\rm prop}}{\left[\prod_{n}F_{n}^{G} \right]_{\rm old}}
\dfrac{\left[\prod_k\text{Normal}\left( h_{\rm prop}^{k};\mu,\upsilon I\right) \right]}{\left[\prod_k\text{Normal}\left( h_{\rm old}^{k};\mu,\upsilon I\right) \right]}
\dfrac{\text{Binomial}\left(B_{\rm prop};K,\gamma/K\right)}{\text{Binomial}\left(B_{\rm old};K,\gamma/K\right)}
\end{align*}
and $A_Q$ is obtained by the ratio of the proposals
\begin{align*}
A_Q
&=
\frac{
Q^{g_1,g_2,\sigma,\gamma}\left(h^{1:K}\old|B\prop,h^{1:K}\prop,B\old\right)
}{
Q^{g_1,g_2,\sigma,\gamma}\left(h^{1:K}\prop|B\old,h^{1:K}\old,B\prop\right)
}
\frac{
Q^{g_1,g_2,\sigma,\gamma}\left(B\old|B\prop,h^{1:K}\prop\right)
}{
Q^{g_1,g_2,\sigma,\gamma}\left(B\prop|B\old,h^{1:K}\old\right)
}
\end{align*}
Because of \emph{assumption~1,} this simplifies to
\begin{align*}
A_Q
&=
\frac{
Q^{g_1,g_2,\sigma,\gamma}\left(h^{1:K}\old|B\prop,h^{1:K}\prop,B\old\right)
}{
Q^{g_1,g_2,\sigma,\gamma}\left(h^{1:K}\prop|B\old,h^{1:K}\old,B\prop\right)
}
\end{align*}
Because of \emph{assumption~2i,} this simplifies further \begin{align*}
A_Q
&=
\frac{
\mathbb{F}^{g_1,g_2,\sigma,\gamma}_{B\prop\to B\old}\left(h^{B\prop}\old|h^{1:K}\prop\right)
}{
\mathbb{F}^{g_1,g_2,\sigma,\gamma}_{B\old\to B\prop}\left(h^{B\old}\prop|h^{1:K}\old\right)
}
\frac{
\mathbb{G}^{g_1,g_2,\sigma,\gamma}_{B\prop\to B\old}\left(h^{B\old}\old|h^{1:K}\prop\right)
}{
\mathbb{G}^{g_1,g_2,\sigma,\gamma}_{B\old\to B\prop}\left(h^{B\prop}\prop|h^{1:K}\old\right)
}
\end{align*}
Because of \emph{assumption~2ii,} this now simplifies to 
\begin{align*}
A_Q
&=
\frac{
\mathbb{G}^{g_1,g_2,\sigma,\gamma}_{B\prop\to B\old}\left(h^{B\old}\old|h^{1:K}\prop\right)
}{
\mathbb{G}^{g_1,g_2,\sigma,1/\gamma}_{B\prop\to B\old}\left(h^{B\old}\prop|h^{1:K}\old\right)
}
\frac{
\mathbb{G}^{g_1,g_2,\sigma,1/\gamma}_{B\old\to B\prop}\left(h^{B\prop}\old|h^{1:K}\prop\right)
}{
\mathbb{G}^{g_1,g_2,\sigma,\gamma}_{B\old\to B\prop}\left(h^{B\prop}\prop|h^{1:K}\old\right)
}
\end{align*}
Because of \emph{assumption~2iii,} this now simplifies to 
\begin{align*}
A_Q
&=
\begin{dcases}
\delta_{R^{g_1,g_2,\sigma,1/\gamma}\left(h^1\old,h^{B\old}\old,R^{g_1,g_2,\sigma,\gamma}\left(h^1\old,h^{B\old}\old,h^{B\old+1}\old\right)\right)}\left(h^{B\old+1}\old\right)
& B\prop=B\old+1
\\
\delta_{R^{g_1,g_2,\sigma,\gamma}\left(h^1\old,h^{B\old-1}\old,R^{g_1,g_2,\sigma,1/\gamma}\left(h^1\old,h^{B\old-1}\old,h^{B\old}\old\right)\right)}\left(h^{B\old}\old\right)
& B\prop=B\old-1
\end{dcases}
\end{align*}
The last equality above comes from \emph{assumption~2i}, which indicates that atoms other than $B\old$ and $B\prop$ do not move.

According to \emph{assumption~2iv,} for the two cases we get:
\begin{align*}
R^{g_1,g_2,\sigma,1/\gamma}\left(h_{\rm old}^1,h^{B_{\rm old}}_{\rm old},
R^{g_1,g_2,\sigma,\gamma}\left(h_{\rm old}^{1},h^{B_{\rm old}}_{\rm old},h^{B_{\rm old}+1}_{\rm old}\right)
\right)
&=
h^{B_{\rm old}+1}_{\rm old},
&
B\prop&=B_{\rm old}+1
\\
R^{g_1,g_2,\sigma,\gamma}\left(h^1_{\rm old},h^{B_{\rm old}-1}_{\rm old},
R^{g_1,g_2,\sigma,1/\gamma}\left(h^1_{\rm old},h^{B_{\rm old}-1}_{\rm old},h^{B_{\rm old}}_{\rm old}\right)
\right)
&=h^{B_{\rm old}}_{\rm old},
& B\prop&=B_{\rm old}-1
\end{align*}
This indicates that the second factor of the acceptance ratio reduces to
\begin{align*}
A_Q
&=
1
\end{align*}
and drops out from the acceptance ratio of our MH. For this reason, the acceptance of the MH sampler reduces to that of the Metropolis sampler \cite{robert2004monte,presse2023data}.


\section{Additional results}
\label{app:add_res}

\begin{figure}[tbp]
\includegraphics[scale=0.7]{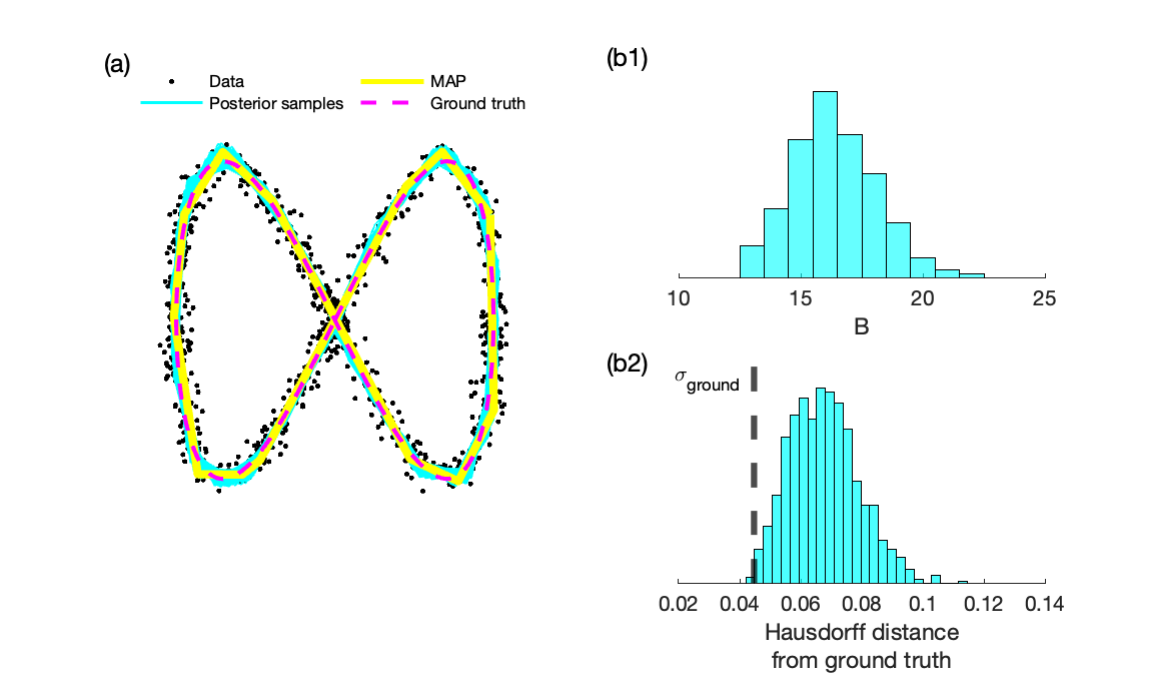}
\centering
\caption{Synthetic data analysis of a Lissajous curve.
Reconstruction is applied on a noisy point-cloud of size $N=1000$ and noise $\tau=500$ (\ie $\sigma\approx0.045$). 
A total of 3,000 MCMC iterations are performed using Sampler~3, with the first 30\% discarded as burn-in and the remaining samples thinned at a ratio of 1:3 for visualization purposes.
The panel on the left shows the point-cloud and reconstructed curves. The panels to the right summarize key posterior results.}
\label{fig:SI_lis}
\end{figure}

\begin{figure}[tbp]
\includegraphics[scale=0.7]{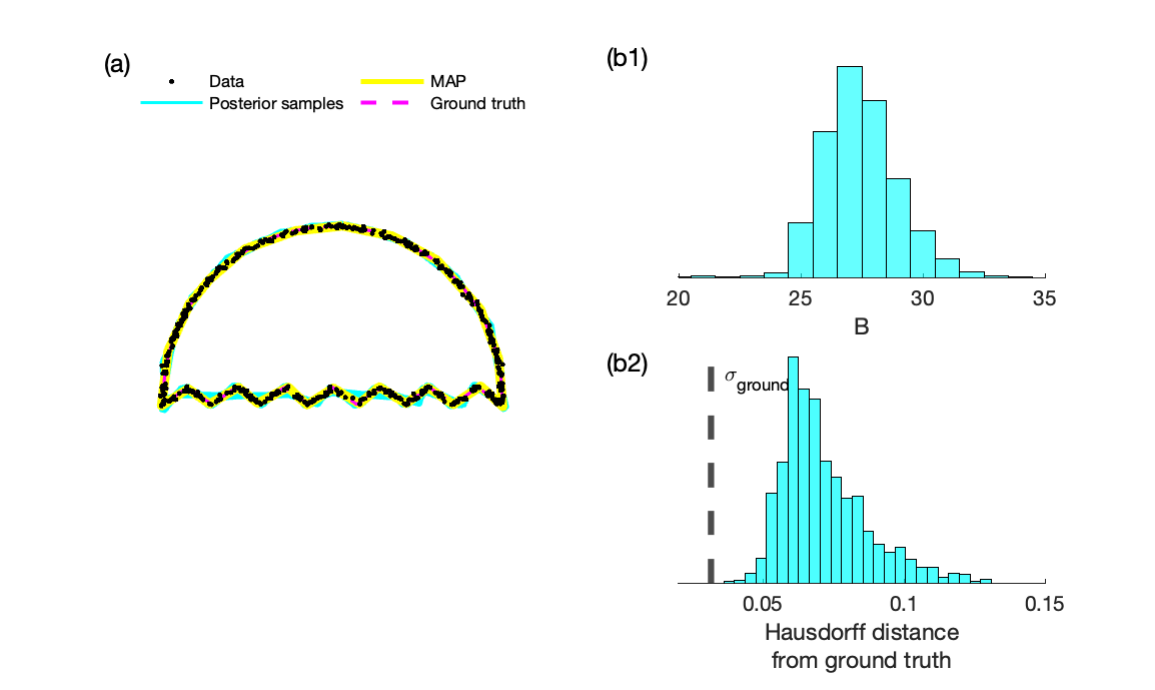}
\centering
\caption{Reconstruction of the \textsc{Sawtooth example}.
A total of 3,000 MCMC iterations are performed using Sampler~3, with the first 20\% discarded as burn-in and the remaining samples thinned at a ratio of 1:3 for visualization purposes.
The panel on the left shows the point-cloud and reconstructed curves. The panels to the right summarize key posterior results.
}
\label{fig:SI_sawtooth}
\end{figure}

In \cref{fig:SI_lis} we evaluate our reconstruction framework on a synthetic dataset generated from an additional Lissajous curve \cite{lawrence2013catalog}, defined by
\begin{align*}
\mathcal{G}=\left\{\left(\sin(t),\sin(2t)\right)\right\}_{t\in[0,2\pi)}\subset\mathbb{R}^2.
\end{align*}
In \cref{fig:SI_sawtooth} we evaluate our reconstruction framework on another synthetic dataset generated from a semicircle with a zigzag lower side. The ground-truth curve is defined by the ordered inducing points
\begin{align*}
\mathcal{H}
=
\left
\{
\left(
\cos\left(\pi-\frac{j\pi}{49}\right),
\sin\left(\pi-\frac{j\pi}{49}\right)
\right)
\right\}_{j=0,1,\ldots,49}
\
\quad\cup
\left\{
\left(
1-\frac{2}{13}\ell,
\frac{(-1)^{\ell+1}}{10}
\right)
\right\}_{\ell=0,1,\ldots,13}
\subset\mathbb{R}^{2},
\end{align*}
where the semicircular inducing points are ordered from left to right and the zigzag inducing points are ordered from right to left, producing a closed counterclockwise curve. This ground-truth curve is chosen to contain both smooth and non-smooth segments and it is similar to the standardized \textsc{Sawtooth example} from Ref.~\cite{ohrhallinger2018stretchdenoise}.

\begin{figure}[tbp]
\includegraphics[scale=0.7]{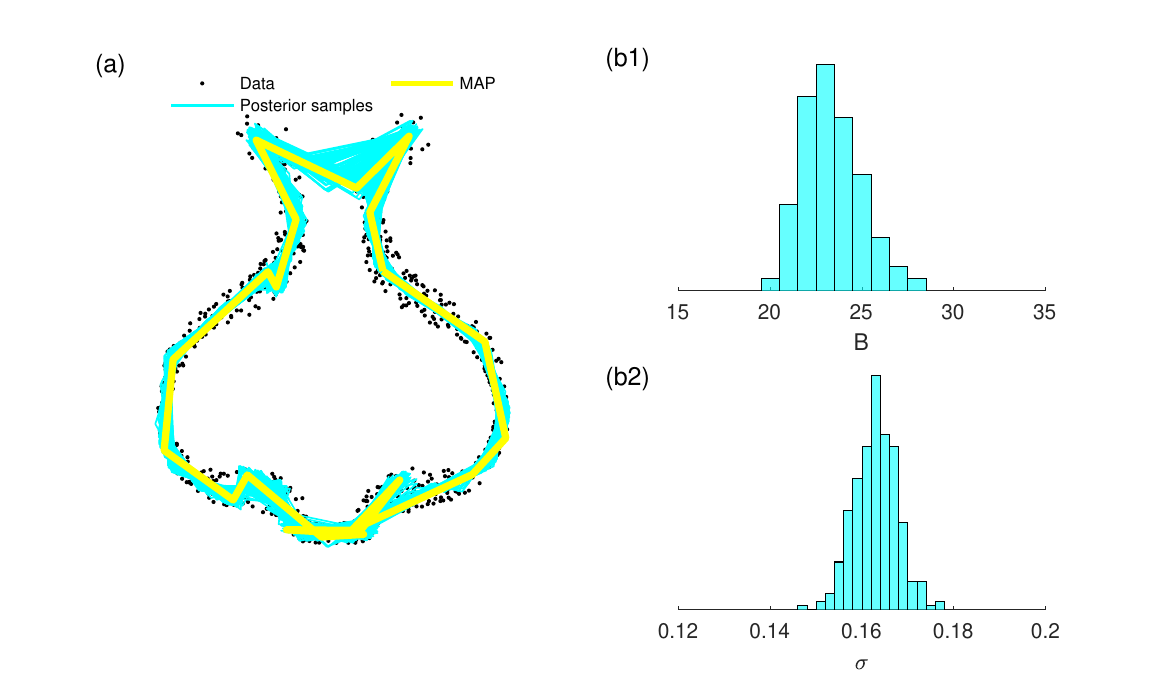}
\centering
\caption{Reconstruction of the \textsc{Bottle example}.
A total of 3,000 MCMC iterations are performed using Sampler~3, with the first 50\% discarded as burn-in and the remaining samples thinned at a ratio of 1:5 for visual purposes.
The panel on the left shows the point-cloud and reconstructed curves. The panels to the right summarize key posterior results.
}
\label{fig:SI_bottle}
\end{figure}

In \cref{fig:SI_bottle} we evaluate our reconstruction framework on the standardized \textsc{Bottle example} from Ref.~\cite{lee2000curve}.

\clearpage

\bibliographystyle{unsrt}
\bibliography{references}


\end{document}

%% file: proposal_diagram.tex
\begin{minipage}{0.49\textwidth}
\centering
\resizebox{\linewidth}{!}{%
\begin{tikzpicture}[x=1.25cm,y=1.10cm]

\path[use as bounding box] (-0.6,-1.2) rectangle (10.0,7.8);


\coordinate (A) at (0.0,3.10);
\coordinate (B) at (0.75,5.15);
\coordinate (C) at (2.55,6.20);
\coordinate (D) at (4.55,4.75);

\coordinate (E) at (4.05,3.15);
\coordinate (F) at (4.55,0.85);

\coordinate (G) at (2.55,0.30);
\coordinate (H) at (0.75,1.65);

\coordinate (R) at (7.05,3.00);


\draw[solidline] (B)--(C)--(D);
\draw[solidline] (B)--(A); 
\draw[solidline] (A)--(H)--(G)--(F);


\draw[solidline] (D)--(F);
\draw[dashedline] (D)--(E);
\draw[dashedline] (F)--(E);


\foreach \P in {A,B,C,D,E,F,G,H,R}
    \node[vertex] at (\P) {};


\draw[arrowline, figorange, arrows={[scale=2]}]
    (R) to[out=155,in=25] (E);

\node[figorange] at (5.65,3.72) {$\mathbb{G}$};


\draw[arrowline, figorange, arrows={[scale=2]}]
    (4.50,0.87)
    .. controls (3.55,0.75) and (3.25,1.55) ..
    (3.68,1.78)
    .. controls (4.25,2.05) and (4.55,1.45) ..
    (4.53,0.98);

\node[figorange] at (3.20,1.70) {$\mathbb{F}$};


\node[align=center] at (3.45,7.20)
{
    $\displaystyle B_{\mathrm{prop}}>B_{\mathrm{old}}$\\[3pt]
    $\displaystyle B_{\mathrm{prop}}=B_{\mathrm{old}}+1$
};

\node[left=7pt] at (E)
{
    $\displaystyle h_{\mathrm{prop}}^{B_{\mathrm{prop}}}$
};

\node[right=10pt,align=left] at (R)
{
    $\displaystyle B_{\mathrm{old}}+1=B_{\mathrm{prop}}$
    \\[6pt]
    $\displaystyle
      h_{\mathrm{old}}^{B_{\mathrm{old}}+1}
      =
      h_{\mathrm{old}}^{B_{\mathrm{prop}}}$
};

\node[below=5pt,align=center] at (F)
{
    $\displaystyle B_{\mathrm{old}}$
    \\[4pt]
    $\displaystyle
      h_{\mathrm{old}}^{B_{\mathrm{old}}}
      =
      h_{\mathrm{prop}}^{B_{\mathrm{old}}}$
};

\end{tikzpicture}%
}
\end{minipage}
\hfill
\begin{minipage}{0.49\textwidth}
\centering
\resizebox{\linewidth}{!}{%
\begin{tikzpicture}[x=1.25cm,y=1.10cm]

\path[use as bounding box] (-0.6,-1.2) rectangle (10.0,7.8);


\coordinate (A) at (0.0,3.10);
\coordinate (B) at (0.75,5.15);
\coordinate (C) at (2.55,6.20);
\coordinate (D) at (4.55,4.75);

\coordinate (E) at (4.05,3.15);
\coordinate (F) at (4.55,0.85);

\coordinate (G) at (2.55,0.30);
\coordinate (H) at (0.75,1.65);

\coordinate (R) at (7.05,3.00);


\draw[solidline] (B)--(C)--(D);
\draw[solidline] (B)--(A); 
\draw[solidline] (A)--(H)--(G)--(F);


\draw[dashedline] (D)--(F);
\draw[solidline] (D)--(E);
\draw[solidline] (F)--(E);


\foreach \P in {A,B,C,D,E,F,G,H,R}
    \node[vertex] at (\P) {};


\draw[arrowline, figorange, arrows={[scale=2]}]
    (E) to[out=15,in=165] (R);

\node[figorange] at (5.65,3.55) {$\mathbb{F}$};


\draw[arrowline, figorange, arrows={[scale=2]}]
    (4.50,0.87)
    .. controls (3.55,0.75) and (3.25,1.55) ..
    (3.68,1.78)
    .. controls (4.25,2.05) and (4.55,1.45) ..
    (4.53,0.98);

\node[figorange] at (3.10,1.68) {$\mathbb{G}$};


\node[align=center] at (3.45,7.20)
{
    $\displaystyle B_{\mathrm{prop}}<B_{\mathrm{old}}$\\[3pt]
    $\displaystyle B_{\mathrm{prop}}=B_{\mathrm{old}}-1$
};

\node[left=8pt,align=center] at (E)
{
    $\displaystyle B_{\mathrm{old}}$
    \\[4pt]
    $\displaystyle h_{\mathrm{old}}^{B_{\mathrm{old}}}$
};

\node[right=10pt,align=left] at (R)
{
    $\displaystyle B_{\mathrm{old}}$
    \\[4pt]
    $\displaystyle h_{\mathrm{prop}}^{B_{\mathrm{old}}}$
};

\node[below=7pt,align=center] at (F)
{
    $\displaystyle B_{\mathrm{old}}-1=B_{\mathrm{prop}}$
    \\[5pt]
    $\displaystyle
      h_{\mathrm{old}}^{B_{\mathrm{prop}}}
      =
      h_{\mathrm{prop}}^{B_{\mathrm{old}}}$
};

\end{tikzpicture}%
}
\end{minipage}